\pdfoutput=1

\documentclass[11pt,a4paper]{article}
\usepackage[hyperref]{acl2020}
\usepackage{times}
\usepackage{latexsym}

\usepackage{microtype}

\aclfinalcopy % Uncomment this line for the final submission
\usepackage{amsmath,amsfonts,bm}

\def\eqref#1{equation~\ref{#1}}
\def\1{\bm{1}}

\DeclareMathAlphabet{\mathsfit}{\encodingdefault}{\sfdefault}{m}{sl}
\SetMathAlphabet{\mathsfit}{bold}{\encodingdefault}{\sfdefault}{bx}{n}

\usepackage{color}
\newcommand{\plotspec}[1]{
    \raisebox{-.5\height}{
        \includegraphics[trim=1.1cm 1.0cm 0.5cm 0.5cm, clip, width=\linewidth, height=1.5cm]{#1}
}}

\usepackage[utf8]{inputenc} % allow utf-8 input
\usepackage[T1]{fontenc}    % use 8-bit T1 fonts
\usepackage{hyperref}       % hyperlinks
\usepackage{url}            % simple URL typesetting
\usepackage{booktabs}       % professional-quality tables
\usepackage{amsfonts}       % blackboard math symbols
\usepackage{nicefrac}       % compact symbols for 1/2, etc.
\usepackage{microtype}      % microtypography
\usepackage{graphicx}
\usepackage{multirow}
\usepackage[inline]{enumitem}
\usepackage{subcaption}
\newcommand{\red}[1]{{\color{red}#1}}

\usepackage{tabularx}
\usepackage{blindtext}

\newcolumntype{L}[1]{>{\raggedright\arraybackslash}p{#1}}
\newcolumntype{C}[1]{>{\centering\arraybackslash}p{#1}}
\newcolumntype{R}[1]{>{\raggedleft\arraybackslash}p{#1}}

\title{Text-Free Image-to-Speech Synthesis Using Learned Segmental Units}

\author{%
  Wei-Ning Hsu$^{1}$\thanks{\hspace{.3em} Equal contribution} \quad 
  David Harwath$^{2*}$ \quad 
  Christopher Song$^3$ \quad 
  James Glass$^1$ \\
  $^1$Massachusetts Institute of Technology \\
  $^2$University of Texas at Austin\\
  $^3$Johns Hopkins University\\
  \texttt{wnhsu@csail.mit.edu}, \texttt{harwath@utexas.edu} \\
}

\begin{document}

\maketitle

\begin{abstract}
In this paper we present the first model for directly synthesizing fluent, natural-sounding spoken audio captions for images that does not require natural language text as an intermediate representation or source of supervision. Instead, we connect the image captioning module and the speech synthesis module with a set of discrete, sub-word speech units that are discovered with a self-supervised visual grounding task. We conduct experiments on the Flickr8k spoken caption dataset in addition to a novel corpus of spoken audio captions collected for the popular MSCOCO dataset, demonstrating that our generated captions also capture diverse visual semantics of the images they describe. We investigate several different intermediate speech representations, and empirically find that the representation must satisfy several important properties to serve as drop-in replacements for text.
\end{abstract}

\begin{figure*}[thb]
    \centering
    \includegraphics[width=.9\linewidth]{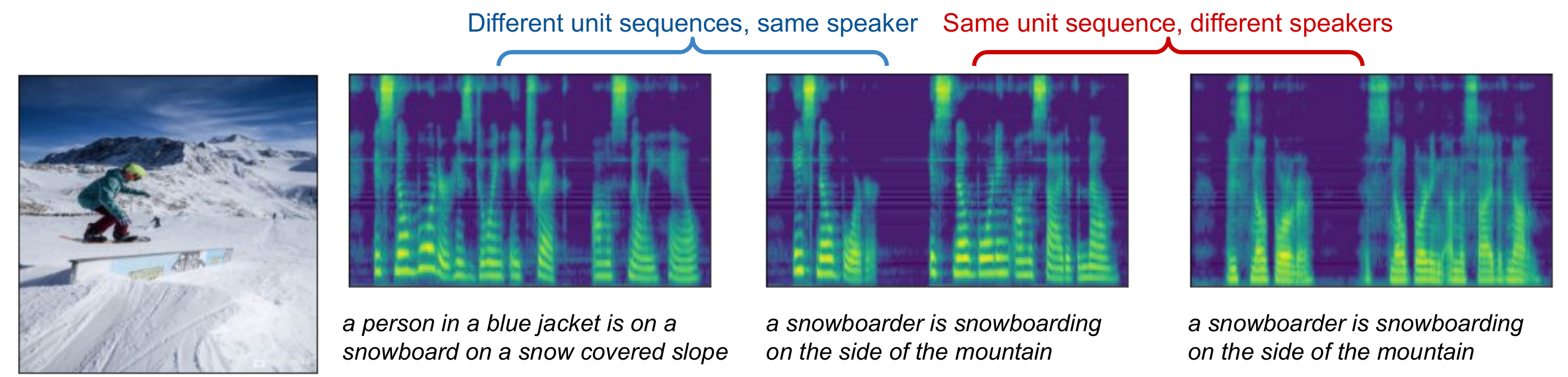}
    \caption{Spoken image captions generated from the proposed model, with diversity in both linguistic content and acoustic properties, controlled through the I2U and the U2S models, respectively. Transcriptions are provided only for illustration. 
    Audio samples are available at \url{https://wnhsu.github.io/image-to-speech-demo}. More samples can be found in Section H and I in the appendix.}
    \label{fig:example}
\end{figure*}

\section{Introduction}
Although there are over 7,000 languages spoken worldwide~\citep{ethnologue}, only several dozen have enough data available to support supervised speech recognition, and many languages do not even employ a writing system~\citep{adda2016breaking}. In contrast, most people learn to use spoken language long before they learn to read and write, suggesting that linguistic annotation is not a prerequisite for speech processing systems. This line of reasoning motivates research that aims to discover meaningful linguistic abstractions (phones, words, etc.) directly from the speech signal, with the intention that they could reduce the reliance of spoken language systems on text transcripts.

A rich body of work has recently emerged investigating representation learning for speech using visual grounding objectives~\citep{synnaeve14,harwath15,harwath16,kamper17a,havard19,merkx19,chrupala17,alishahi17,scharenborg18,hsu2018disentangling,kamper19,suris19,ilharco19,eloff19}, as well as how word-like and subword-like linguistic units can be made to emerge within these models~\citep{harwath17,harwath19,drexler17,alishahi17,harwath19,harwath19a,havard19a,harwath_hsu_glass_20}. So far, these efforts have predominantly focused on \textit{inference}, where the goal is to learn a mapping from speech waveforms to a semantic embedding space. \textit{Generation} of speech conditioned on a point in a semantic space has been less explored, and is what we focus on in this work. We hypothesize that generative approaches offer interesting advantages over relying solely on inference. For example, prior works have demonstrated the capability of recognizing visually descriptive words, but have not been shown to learn non-visual words or grammar. Our experiments show that these aspects of spoken language are learned to some degree by a visually-grounded generative model of speech.

Specifically, we introduce a model capable of directly generating fluent spoken audio captions of images without the need for natural language text, either as an intermediate representation or a form of supervision during training (Figure~\ref{fig:example}). Tremendous progress has been made recently in natural language image caption generation~\citep{kiros2014multimodal,mao2015deep, vinyals2015show,karpathy2015deep,xu2015show,rennie2017self,dai2017contrastive,lu2017knowing,anderson2018bottom,lu2018neural} and naturalistic text-to-speech synthesis (TTS)~\citep{ping2017deep,taigman2017voiceloop,tacotron,tacotron2,wavenet}. 
Combining these models provides a means for generating spoken image descriptions, but existing approaches for training these models are reliant on text during training. Instead, we leverage sub-word speech units discovered using a self-supervised learning objective as a drop-in replacement for the text. We hypothesize that by using such techniques, an even wider variety of traditionally text-based NLP models could be applied to speech data without the need for transcription or automatic speech recognition (ASR) systems. Because all human languages utilize small, discrete phonetic inventories~\citep{international1999handbook}, we posit that our framework should be applicable for any language in the world. In our experiments, we demonstrate that not just any set of discovered speech units can function in this role. We find the greatest success with units that are \emph{discrete}, exhibit a \emph{low frame-rate}, and \emph{highly robust} to speaker and environmental variability. The main contributions of our paper are as follows:

\textbf{1. The first methodology for fluent image-to-speech synthesis that does not rely on text.} A critical aspect of our approach is factorizing the model into an Image-to-Unit (I2U) module and a Unit-to-Speech (U2S) module, where the speech units are discovered in a self-supervised fashion. This approach enables disentanglement of linguistic variability and acoustic/speaker variability.

\textbf{2. Extensive analysis on the properties required for learned units to replace text.} While the idea may seem simple and straightforward, obtaining proper units is not a trivial task. In fact, most of the units experimented in this paper fail to serve as drop-in replacements. Moreover, we demonstrate that what are deemed good units vary significantly for inference and generation.

\textbf{3. Demonstrating insufficiency of beam search-based evaluation.} We show that even when an I2U model fails to generate sensible caption through beam search decoding, it can still produce reasonable captions by sampling from the posterior, hinting that posterior mode-based evaluation can only inspect limited aspects of a model.

\textbf{4. Proposing a semantic diversity-aware metric.} We identify issues of an existing metric~\citep{vijayakumar2018diverse} and propose M-SPICE for sampling-based evaluation to address the problems.

\textbf{5. Over 600,000 spoken audio captions for the MSCOCO dataset.} We collect 742 hours of speech from 2,352 people tasked with reading each caption out loud. This dataset will be made publicly available to support work at the intersection of speech, language, and vision.

\section{Related Work}
\label{sec:related_work}

\textbf{Image-to-Text and Image-to-Speech Captioning.} Significant progress towards generating realistic (text) captions that describe the content of visual images was made with the advent of deep neural networks~\citep{vinyals2015show,karpathy2015deep,xu2015show,anderson2018bottom}. Far less work has focused on generating spoken audio captions from natural images. Training an image-to-speech system using separate $(image, text)$ and $(text, speech)$ datasets was explored in~\citep{ma2019unpaired}. \citet{hasegawa2017image2speech} is the only prior work that has explored image-to-speech synthesis without using text~\citep{hasegawa2017image2speech}, but with limited results. In that work, BLEU scores were only computed in terms of unsupervised acoustic units, not an estimate of the actual words produced by the synthesizer, which can be problematic as discussed in Section~\ref{sec:experiments}. The resulting captions were not evaluated for fluency, naturalness, or intelligibility, and the BLEU scores in terms of the unsupervised units were very low (0.014 on the MSCOCO test set) compared to ours (0.274). 
\citet{wang2020show} is a concurrent work that proposes a text-free end-to-end image-to-speech model, which simplifies the task by using pairs of image and synthesized speech generated from a single-speaker TTS model to reduce the acoustic variation. In contrast, by leveraging robust learned units, the I2U module in our system can be trained on speech with abundant variation, while the U2S module serves as a vocoder and only requires a small amount of clean speech (transcripts not needed), which imposes less constraints on the data and still outperforms \citet{wang2020show} on captioning metrics.

\textbf{Speech Synthesis without Text and Voice Conversion.} Voice conversion is a classic problem in speech processing that involves resynthesizing a recording to sound as if it was spoken by a different person~\citep{abe1990voice,stylianou1998continuous,toda2007voice}. Voice conversion has recently seen progress using neural approaches~\citep{hsu2016voice,hsu2017learning,hsu17a,hsu2017voice,fang2018high,chorowski2018using,chou2018multi,lorenzo2018voice,stargan,blow}, but the most relevant work to our own is the ZeroSpeech 2019 challenge~\citep{zerospeech19, tjandra2019vqvae, cho19}, which addresses unsupervised learning of discrete speech units that can be used as the basis of a synthesizer. Unlike image-to-speech synthesis, these tasks only infer units from given audio recordings and do not require generation.

\textbf{Speech Pre-Training for Downstream Tasks.} Interest in un/self-supervised pre-training has recently surged in the speech community. Several papers have investigated masked prediction~\citep{Baevski2020vq-wav2vec, Schneider2019a, wang2020unsupervised}, while others have implemented autoencoder models with various constraints applied to the latent representations, such as discreteness~\citep{vqvae,eloff19,chorowski19} or disentanglement~\citep{hsu17a,hsu2018scalable,khurana2019factorial}. Predictive coding has recently seen a revival using deep neural networks~\citep{cpc,chung19}. Most work applied pre-trained speech models to only inference problems such as ASR~\cite{baevski2019effectiveness} or phone discrimination~\cite{kharitonov2020data}, with a notable exception being~\citet{tjandra2019speech}, which focuses on text-free machine translation. 

\begin{figure*}[t!]
    \centering
    \includegraphics[width=\linewidth]{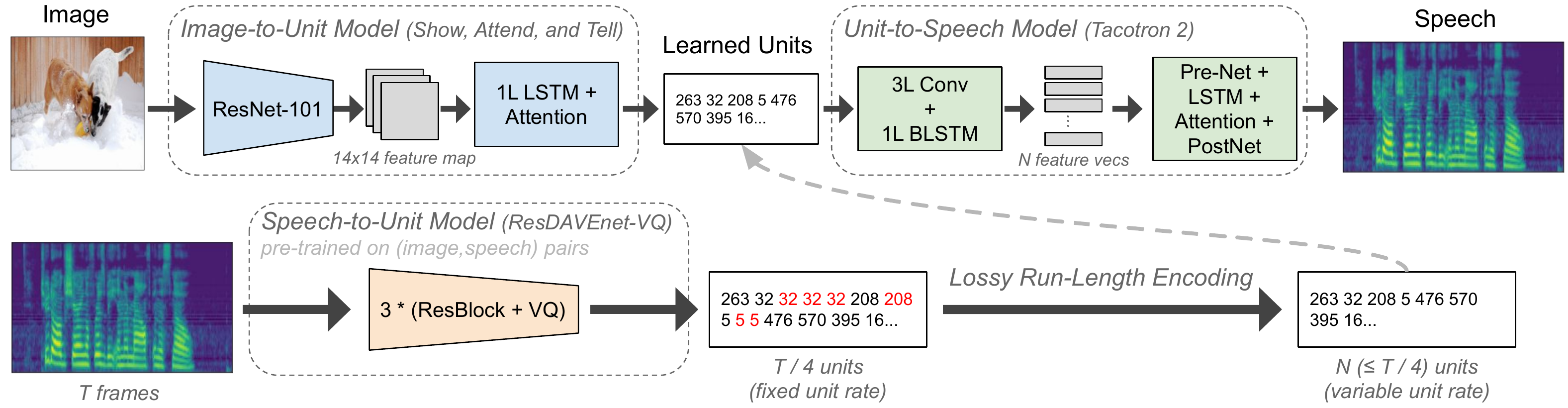}
    \caption{Diagram of our proposed framework. The ResDAVEnet-VQ model was trained using a $\{2\}\rightarrow\{2,3\}$ curriculum (in the notation given in \citet{harwath_hsu_glass_20}).}
    \label{fig:diagram}
\end{figure*}

\begin{table*}
    \centering
    \small
    % \resizebox{\linewidth}{!}{
    \begin{tabular}{llrrrrr}
    \toprule
    & \textbf{Data} & \textbf{Hr} & \textbf{\#Utt} & \textbf{\#Spk} & \textbf{Maj. Spk} & \textbf{Description} \\ \midrule
    S2U & PlacesAudio~\citep{harwath16} & 936 & 400K & 2683 & American & spontaneous image caption \\ \midrule
    \multirow{2}{*}{I2U}
    & Flikr8kAudio~\citep{harwath15} & 46 & 40K & 183 & \multirow{2}{*}{American} & \multirow{2}{*}{scripted image caption} \\
    & SpokenCOCO (this work) & 742 & 605K & 2353 & \\ \midrule
    \multirow{2}{*}{U2S}
    & LJSpeech~\citep{ljspeech17} & 24 & 13K & 1 & American & read non-fiction books \\
    & VCTK~\citep{vctk} & 44 & 40K & 109 & British & read newspaper \\
    \bottomrule
    \end{tabular}
    % }
    \caption{\label{tab:data} Speech dataset summary. For training S2U and I2U models, their corresponding image datasets, MSCOCO~\citep{mscoco}, Flickr8k~\citep{flickr8k}, and Places~\citep{places}, are also used.}
\end{table*}
\section{Method}

\subsection{Framework Overview}
A depiction of our modeling approach is shown in Figure~\ref{fig:diagram}. Caption generation for an image involves a cascade of two components: given an input image $I$, we first generate a linguistic unit sequence $U$ according to the I2U module $P(U \mid I)$. Given the linguistic symbol sequence $U$, we generate a speech waveform $S$ according to the U2S module $P(S \mid U)$. If the linguistic unit sequence $U$ were to take the form of natural language text, the model would be equivalent to the cascade of a conventional image captioning system followed by a TTS module. Note that we assume $S \perp I \mid U$ because prosody variation is not dependent on the image for the datasets considered.

The key idea in this paper is to instead define $U$ to be a sequence of \emph{learned} speech units that are as \textit{robust and compact} as possible like text, but discovered without text supervision. 
% Specifically, we use the visually-grounded sub-word speech unit discovery model introduced by~\cite{harwath_hsu_glass_20}. 
We define inference with this S2U model as $U = f(S)$, enabling us to ``transcribe'' any given speech audio waveform $S$ into a sequence of units $U$. The addition of this third component enables us to train $P(U \mid I)$ from a dataset of images paired with spoken captions $\{(I_1, S_1),\dots,(I_N, S_N)\}$. The conditional independence assumption between $S$ and $I$ given the $U$ enables us to choose any arbitrary speech dataset for training $P(S \mid U)$, therefore enabling the speaker characteristics and other acoustic properties to be independently controllable from the I2U system~\citep{wang2018style,hsu2018hierarchical,henter2018deep,akuzawa2018expressive}.
% ~\citep{wang2018style,hsu2018hierarchical,skerry2018towards,henter2018deep,akuzawa2018expressive,zhang2019learning,habib2019semi}.

\subsection{Datasets}
Table~\ref{tab:data} summarizes the five datasets used for training S2U, I2U, and U2S models. Note that we deliberately choose different datasets for training each module, which aims to examine the robustness of the units when transferring across domains, including shift in speaker demography, speaking style (scripted/spontaneous), and linguistic content (book/newspaper/image description).

Specifically, as part of this work we collect \textbf{SpokenCOCO}, a spoken version of the MSCOCO captioning dataset~\citep{mscoco}, via Amazon Mechanical Turk by displaying the text to a person and having them read it aloud. Additional details regarding the dataset statistics can be found in Section A in the appendix. 
Note that although there exists a speech version of MSCOCO named Speech-COCO~\cite{havard2017speech}, it is comprised of only synthesized speech using a concatenative TTS model in eight speakers' voice. Disfluencies (e.g., ``uh'') are randomly inserted to imitate real speech. As a result, it offers limited diversity and naturalness compared to SpokenCOCO presented here.

\subsection{Learning Robust Linguistic Units from Visually-Grounded Speech}
\label{ssec:units}
We propose to build the S2U model upon ResDAVEnet-VQ, an audio-visual grounding model introduced in~\citet{harwath_hsu_glass_20} that has shown to learn discrete phone-like and word-like units in the intermediate vector quantizing (VQ) layers. The ResDAVEnet-VQ model is trained to associate speech with contextually relevant visual inputs using a triplet loss~\citep{weinberger09}, which can be interpreted as maximizing a lower bound on the mutual information between image and speech~\citep{Tschannen2020On}. Since visual semantics are described with words, which in turn are composed of phones, the intermediate representations learned by ResDAVEnet-VQ are forced to be predictive of words and phones rather than speaker identity, acoustic environment, noise, etc. 
On the other hand, the majority of un/self-supervised speech representations are trained by reconstructing~\citep{chorowski19,hsu17a} or predicting unseen speech signals~\citep{chung19}, which would inevitable model factors unrelated to the linguistic content. To demonstrate the advantage of models trained with grounding based objectives, we will compare ResDAVEnet-VQ with a reconstruction based model, WaveNet-VQ~\cite{cho19}, trained on the PlacesAudio dataset. Note that there are more recent studies learning speech representations without reconstructing speech~\cite{Baevski2020vq-wav2vec}, which may enjoy the same benefit as grounding models. We will leave the study for future work.

\subsection{Unit Selection and Run Length Encoding}
Although the ResDAVEnet-VQ model has been shown to be capable of learning a hierarchy of discrete speech units from phone-like to word-like, the experiments in~\citep{harwath_hsu_glass_20} show that only several hundred words are explicitly learned, which tend to be ``visual words.'' Conversely, the phone-like units learned by the lower VQ layers of the model were shown to cover all of the phones in American English (as there are only several dozen). For this reason, we choose to use phone-like units learned by the lower VQ layers to represent $U$. We postulate that future work may benefit from incorporating the word-level units. 

Nominally, the VQ layers will output one-hot vectors at a uniform temporal rate, which is downsampled with respect to the framerate of the acoustic input depending upon which VQ layer is used. Given an input spectrogram to the model computed with a 10ms frame shift, the two VQ layers we investigate in this paper (VQ2 and VQ3) respectively output vectors every 20ms and 40ms. In general, the VQ units from each model are repeated for several consecutive frames. We can decrease the average length of the symbol sequence $U$ by employing a lossy form of \textbf{run-length encoding} (RLE) (see Figure~\ref{fig:diagram}) which retains the sequence of symbol identities but discards duration information. This removes the burden of unit duration modeling from the I2U model and shifts it onto the U2S model, which we will show to be crucial.

\subsection{Image-to-Unit and Unit-to-Speech}
Both the I2U model and the U2S model are based upon recurrent seq2seq with attention networks~\citep{bahdanau2014neural}. Specifically, we use an off-the-shelf implementation of Show-Attend-and-Tell (SAT)~\cite{xu2015show} for the I2U model. It has an image encoder pre-trained for classification, which is language agnostic and hence should work in any language within our proposed framework. We choose this model not only because of its ubiquity, but also because it represents one of the simplest captioning models by today's standards. After inferring the discrete unit sequence $U$ for each speech recording $S$ in the training dataset using the ResDAVEnet-VQ model, we train the I2U model on the resulting $(I, U)$ pairs of images and their unit sequences representing the caption. 

For the U2S model, we use the off-the-shelf implementation of Tacotron-2~\cite{tacotron2} with no pre-training to generate spectrograms, and WaveGlow~\citep{waveglow} to generate waveforms from these spectrograms. TTS models generally require training pairs of the form $(text, speech)$, but in our case the ResDAVEnet-VQ unit transcription model enables us to train the Tacotron-2 model on any arbitrary set of speech waveforms by transcribing each recording $S$ into its unit sequence $U$, and then training the model on the $(U, S)$ pairs. Both models are trained with the maximum likelihood objective, with the I2U model trained to maximize $\mathbb{E}_{I, U} \left[ \log P(U \mid I)\right]$ and the U2S model trained to maximize $\mathbb{E}_{S, U} \left[ \log P(S \mid U)\right]$.

\section{Experiments}
\label{sec:experiments}
We design experiments to address three questions:

\textbf{First, how can we measure the performance of an image-to-speech system?} Generating a spoken caption for an image within our framework first involves generating a latent unit sequence with the I2U model, followed by synthesizing a waveform conditioned on the unit sequence with the U2S model. The concatenation of these models can fail to produce a good caption if the I2U model fails to encode linguistic/semantic information into the unit sequence, or if the U2S model fails to synthesize an intelligible waveform given a unit sequence. 
To better localize these failure modes, we evaluate the full I2S system as well as the U2S system in isolation. We evaluate the U2S system by using it as a vocoder to synthesize unit sequences \textit{inferred from real speech} and soliciting human judgements in the form of Mean Opinion Score (MOS) and Side-By-Side (SXS) preference tests. 

To evaluate the full I2S system, we could use any method that measures the semantic information contained in the generated language. One way would be to use a speech-to-image retrieval system, while another is to infer a text transcription of the generated speech with an ASR system, and then employ standard text-based caption evaluation metrics. This has the advantages of making possible a comparison with an "upperbound" system in the form of text-based captioning, as well as measuring other aspects of language generation such as syntactic correctness (via BLEU-4) and scope of the learned vocabulary. For those reasons, we use the latter strategy.
One may consider adopting unit-based metrics, such as BLEU-4 on units. However, there are two main caveats: first, systems built on different units are not directly comparable; second, scores can be inflated if duration is modeled as repeating units. More discussions can be found in Section D in the appendix.

\textbf{Second, what properties must learned units have in order to be a drop-in replacement for text?} The most essential differences between text and speech are the amount of information encoded and the sequence lengths. Beyond text, speech also encodes prosody, speaker, environment information as well as the duration for each linguistic unit, all of which are minimally correlated with the conditioned images. We hypothesize that learned speech units should discard such information in order to seamlessly connect the I2U and U2S modules.
To verify it, we pay particular attention to the variations of the learned units in \textit{frame rate} (VQ2 or VQ3), \textit{encoding of duration information} (using RLE or not), and \textit{robustness to domain shift} (WaveNet-VQ or ResDAVEnet-VQ). The units extracted from the WaveNet-VQ model are denoted as WVQ, and units are run-length encoded by default. Table~\ref{tab:mos_wer} shows the properties of the units, where it is shown that VQ3 and WVQ have the same framerate, and VQ3 and VQ2 perform similarly well on phone discrimination (ABX error, the lower the better), which is an inference task.

\textbf{Third, how should language generation models be evaluated more generally?} We examine evaluation of the I2S model using beam search-based decoding as well as sampling-based decoding. We find that because evaluation metrics that are reliant on beam search-based decoding only evaluate the \textit{mode} of a model's posterior, they do not reflect the ability of a model to generate diverse linguistic content. Furthermore, we show that it is possible for a model's posterior mode to be linguistically meaningless, and yet meaningful language can still be generated with sampling-based decoding. Towards this end, we introduce a novel multi-candidate evaluation metric (M-SPICE).

\begin{table}[htb]
    \begin{subtable}{\linewidth}
    \centering

    \resizebox{\linewidth}{!}{
    % \small
    \begin{tabular}{c|cc|cc}
        \toprule
        \multirow{2}{*}{Unit} & ABX & Frame
        & \multicolumn{2}{c}{MOS} \\
        & Error & Rate & LJSpeech & SpokenCOCO \\
        \midrule\midrule
VQ3 & 14.52\% & 40ms & 3.723 $\pm$ 0.039 & 3.336 $\pm$ 0.044 \\
VQ2 & 12.51\% & 20ms & 3.932 $\pm$ 0.036 & 2.961 $\pm$ 0.045 \\
WVQ & 24.87\% & 40ms & 3.658 $\pm$ 0.040 & 2.896 $\pm$ 0.053 \\
        \bottomrule
        \end{tabular}
    }
    \caption{\label{tab:mos_wer} Properties of the units and MOS of the U2S models trained on these units with 95\% confidence interval. ABX errors are computed on the ZeroSpeech 2020 English test set.}
    \end{subtable}
    
    \vspace{1em}
    
    \centering
    \begin{subtable}{\linewidth}
    \resizebox{\linewidth}{!}{
    \begin{tabular}{cc|ccc|ccc}
        \toprule
        \multicolumn{2}{c|}{Unit} & 
        \multicolumn{3}{c|}{LJSpeech} & \multicolumn{3}{c}{SpokenCOCO} \\
        A & B & A & Same & B & A & Same & B \\
        \midrule\midrule
VQ3 & VQ2 & 23.9 & 31.5 & 44.6 & 40.4 & 32.5 & 27.5 \\
VQ3 & WVQ & 36.6 & 37.1 & 26.3 & 58.3 & 21.8 & 19.9 \\
        \bottomrule
    \end{tabular}
    }
    \caption{\label{tab:sxs} SXS preference (\%) of the U2S models.}
    \end{subtable}
    
    \caption{\label{tab:mos_sxs}Subjective evaluation of U2S models trained on LJSpeech and re-synthesize units inferred from LJSpeech or SpokenCOCO recordings.}
\end{table}

\subsection{Evaluating the U2S Model}
We construct a Tacotron-2 model for each of the three unit types on the LJSpeech audio data by transcribing each LJSpeech utterances into an unit sequence, then train the U2S model from the run-length encoded unit sequence and waveform pairs. We evaluate the naturalness of the speech produced by each model on held-out data, both in-domain using LJSpeech and out-of-domain using SpokenCOCO. Amazon Mechanical Turk (AMT) workers performed Side-by-Side preference tests (SXS) and naturalness evaluation based on mean opinion scores (MOS) on a scale from 1 to 5 for each U2S model, which we display in Table~\ref{tab:mos_sxs}. Although VQ2 was preferred for in-domain synthesis on LJSpeech, VQ3 achieved the highest scores and least degradation (-0.387) on the out-of-domain SpokenCOCO, indicating that out of the three units VQ3 has the strongest robustness to domain shift.

\begin{table*}[thb]
    \centering
    \small
    \begin{tabular}{cc|ccccc|ccccc}
        \toprule
        \multirow{2}{*}{model} & \multirow{2}{*}{$U$} & \multicolumn{5}{c|}{MSCOCO} & \multicolumn{5}{c}{Flickr8k} \\ 
        & & B-4 & M & R & C & S & B-4 & M & R & C & S \\
        \midrule\midrule
\citet{xu2015show}           & word & 0.243 & 0.239 & -     & -     & - & 0.213 & 0.203 & -     & -     & - \\
\citet{lu2017knowing}        & word & 0.327 & 0.260 & 0.540 & 1.042 & - & - & - & - & - & - \\
\midrule
\citet{wang2020show}         & N/A  & -     & -     & -     & -     & - & 0.035 & 0.113 & 0.232 & 0.080 & - \\
        \midrule
\multirow{3}{*}{SAT}
& word     & 0.315 &0.253 & 0.533 & 0.984 & 0.185   &0.216 & 0.207 & 0.469 & 0.550 & 0.149 \\
& char     & 0.289 &0.239 & 0.512 & 0.879 & 0.172   &0.190 & 0.190 & 0.441 & 0.476 & 0.136 \\
& VQ3      & 0.186 &0.186 & 0.446 & 0.584 & 0.127   &0.116 & 0.141 & 0.390 & 0.232 & 0.091 \\
        \midrule
\multirow{3}{*}{SAT-FT}
& word     & 0.339 & 0.265 & 0.551 & 1.062 & 0.196  & 0.225 & 0.215 & 0.483 & 0.584 & 0.155 \\
& char     & 0.323 & 0.256 & 0.536 & 1.002 & 0.187  & 0.191 & 0.196 & 0.450 & 0.519 & 0.143 \\
& VQ3      & 0.233 & 0.212 & 0.478 & 0.732 & 0.149  & 0.125 & 0.145 & 0.391 & 0.245 & 0.095 \\
        \bottomrule
    \end{tabular}
    \caption{\label{tab:main} Word-based caption evaluation using \textbf{B}LEU\textbf{-4}, \textbf{M}ETEOR, \textbf{R}OUGE, \textbf{C}IDEr, and \textbf{S}PICE. ASR is used to transcribe the spoken captions generated by the proposed VQ3 model into text for evaluation. The beam size $\in \{3, 5, 8, 10\}$ was chosen for each model to maximize SPICE. Our word-based SAT models outperform~\citep{xu2015show} because we use a stronger image encoder (ResNet-101). Full results with all units are in Section E.}
\end{table*}
\subsection{Incorporating the I2U Model}
We trained an SAT model on SpokenCOCO for each of the three run-length encoded units, as well as VQ3 units without RLE. We also compare to text characters and words; the full hyperparameter and training details for all models are provided in Section B in the appendix, but in general we kept these as constant as possible when comparing different linguistic representations. 

Before connecting the U2S model, we noticed that all RLE speech unit models except the one trained on VQ3 units failed during \textit{beam search} decoding on the test images (WVQ consistently failed, while VQ2 sometimes succeeded); rather than producing a diverse sequence of output units, the decoder would generally get stuck in a loop until the maximum decoding length was reached. This phenomenon also took place using VQ3 units without RLE, indicating that the decoder had difficulty modeling the duration of the units. Example outputs are provided in Table A4 in the appendix.
We hypothesize that the reason the VQ2 and WVQ units failed is due to their lack of invariance to domain shift, as evidenced by their decay in naturalness when used for out-of-domain synthesis as shown in Table~\ref{tab:mos_sxs}. This may cause the entropy of the unit distribution conditioned on an image to be higher (each phoneme may be represented by multiple units), and therefore the image-conditioned language model of units suffers from the same looping issues as the unconditional language model of text, as observed in~\citep{holtzman2018learning,fan2018hierarchical,radford2019language,holtzman2019curious,kulikov2019importance, welleck2019neural}.

To evaluate the full Image-to-Speech model, we first train an ASR system on the re-synthesized SpokenCOCO captions using the VQ3 Tacotron-2 model. 
% Specifically, we use the Wav2Letter++~\citep{pratap19} implementation of the seq2seq+TDS model~\citep{hannun2019sequencetosequence}, using the COCO caption text as the ground truth labels for ASR training. 
This enables us to estimate a word-level transcription of the spoken captions produced by our system, which we can then use to compute standard captioning evaluation metrics (including BLEU-4~\cite{bleu}, METEOR~\cite{meteor}, ROUGE-L~\cite{rouge}, CIDEr~\cite{cider}, and SPICE~\cite{spice}) and make direct comparisons with conventional text captioning systems. 
In order to verify that the synthesized captions are intelligible to humans and the ASR system did not simply learn to recognize artifacts of the synthesized speech, we asked AMT workers to transcribe into words a set of 500 captions generated by our system and also evaluated their naturalness. 
3 workers transcribed and 3 workers rated each caption, allowing us to compute an \textbf{MOS score (3.615$\pm$0.038)}, a word error rate (\textbf{WER}) between the 3 human transcriptions (9.40\%), as well as an average WER between the human and the ASR-produced transcriptions (\textbf{13.97\%}). This confirms that our system produces reasonably natural speech and ASR is sufficiently accurate for transcribing synthesized speech for text-based evaluation.

Table~\ref{tab:main} summarizes our results on MSCOCO and Flickr8k using beam search. Results for VQ2, WVQ, and VQ3 without RLE are omitted because they are all close to zero, but the full table can be found in Section E in the appendix. Unit-based evaluation are presented in Section D for completeness. We compare with the literature for bottom-up text captioning (row 1-2) and text-free end-to-end image-to-speech (row 3). We train the decoder of an SAT model while keeping the image encoder fixed (row 4-6), in addition to fine-tuning the image encoder (row 7-9). 
Despite having no access to text, \textit{the SAT-FT speech captioning model trained on VQ3 units achieves a BLEU-4 score of .233 with beam search decoding on MSCOCO. This is very close to the .243 achieved by the original SAT word-based captioning model.} Figure~\ref{fig:example} shows that the generated captions are fluent and reflect the implicit learning of some syntactic rules. It is evident that the proposed model is capable of generating fluent and meaningful image captions.

\begin{figure*}[htb]
    \centering
    \includegraphics[width=.95\linewidth]{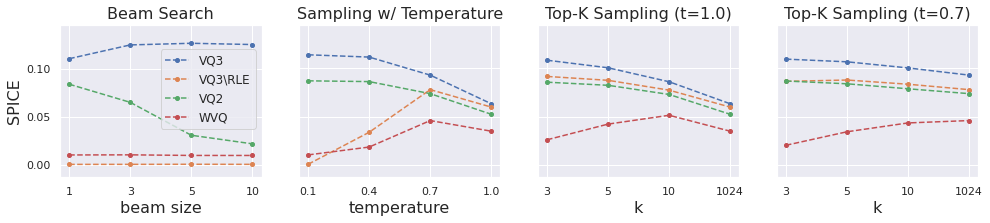}
    \caption{MSCOCO test set SPICE of various units and decoding methods. VQ3\textbackslash RLE denotes VQ3 units without RLE. Top-$k$ sampling considers the $k$-most probable units at each step. Other metrics are shown in Figure A3.}
    \label{fig:unit_comp}
\end{figure*}

\begin{figure}[htb]
    \centering
    \includegraphics[width=.95\linewidth, trim=0 0 17cm 0, clip]{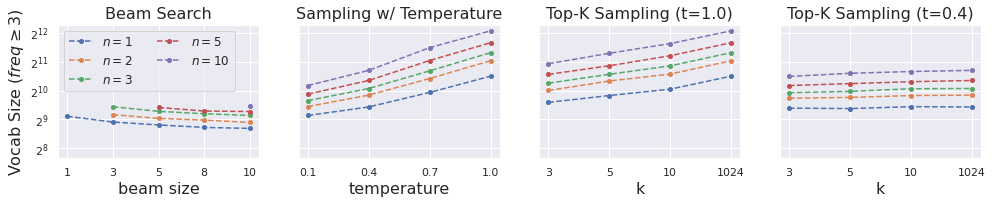}
    \caption{Vocabulary size learned by the proposed I2S model (on MSCOCO). Full results are in Section G.}
    \label{fig:vocab}
    
    \vspace{1em}
    
    \includegraphics[width=.95\linewidth, trim=0 7.5cm 17cm 0, clip]{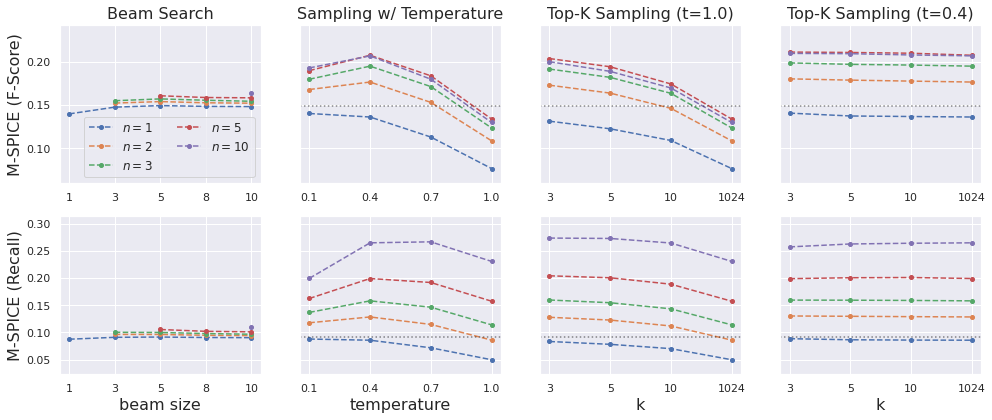}
    \includegraphics[width=.95\linewidth, trim=0 0 17cm 14cm, clip]{figs/mspice_stage2_lg.png}
    \caption{M-SPICE score on MSCOCO. Black dashed lines show the highest value reported for beam search when n=1. M-SPICE recalls are shown in Figure A2.
    }
    \label{fig:mspice_s2}
\end{figure}

\subsection{From Mode to Distribution: Sampling-Based Caption Evaluation}
\label{ssec:sampling}
The results in the previous section only evaluate beam search decoding with the I2U model, and do not fully reveal the posterior over captions for an input image, or whether the unit representations that failed with beam search would work well with other methods. To probe this, we evaluate the models using sampling-based caption generation. Figure~\ref{fig:unit_comp} shows the SPICE scores on SpokenCOCO using beam search and two sampling-based methods. VQ3 still performs the best of all unit types with both beam search and sampled decoding. VQ2 can sometimes generate captions with beam search when the beam is kept small, but as the beam grows it begins to loop and the scores become very low. \textit{We see that all unit types can generate reasonable captions when using sampling-based decoding}, although ResDAVEnet-VQ units consistently outperform the WaveNet-VQ units, suggesting that they better capture sub-word structure. Example outputs are provided in the appendix.

We estimated the vocabulary size of our SAT-FT model using VQ3 units by counting the number of unique recognized words produced at least 3 times when captioning the SpokenCOCO test images. These numbers are shown for the model under the various decoding methods in Figure~\ref{fig:vocab}. The number of captions per image is denoted by $n$, where top candidates are used for beam search and i.i.d. samples are drawn for sampling. Sampling-based decoding reveals a larger vocabulary size than beam search, and the number of words learned by our models ($\ge 2^{12}$) is far greater than the number of words learned by the ResDAVEnet-VQ model (approx. 279) in~\citep{harwath_hsu_glass_20}. We hypothesize that training a model to \textit{generate} spoken captions encourages it to learn many more words than only being trained to retrieve images from captions. We also hypothesize that because beam search attempts to find the mode of the posterior distribution over captions, it tends to produce a smaller set of common words and does not reveal the breadth of the model distribution.

\subsection{New Diversity-Aware Metric: M-SPICE}
The previous section showed that even when the SPICE scores were comparable, sampling-based decoding revealed a much larger model vocabulary than beam search, especially when multiple captions are generated for each image. This highlights a limitation of SPICE in measuring the \textit{diversity}. Formally speaking, SPICE computes an F-score between two bags of semantic propositions $T(S)$ and $T(c)$ parsed from a set of references $S=\{s_i\}_i$ and a hypothesis $c$, where $T(c)$ denotes a bag of propositions extracted from a scene graph parsed $c$, and we can compute that for multiple sentences with $T(S)=\cup_i(T(s_i))$. 

To extend SPICE for scoring multiple hypotheses $C=\{c_j\}_{j=1}^J$, one can compute an average SPICE: $\frac{1}{J} \sum_j F1(T(S), T(c_j))$, or use the oracle SPICE proposed in \citet{vijayakumar2018diverse}: $max_j F1(T(S), T(c_j))$. However, these metrics fail to capture the diversity \textit{among} hypotheses. Consider two hypothesis set, $C^1 = \{c^1_1, c^1_2\}$ and $C^2 = \{c^2_1, c^2_2\}$, where $T(c^1_1)=T(c^1_2)=T(c^2_1)=$ \{(girl), (table), (girl, sit-at, table)\}, $T(c^2_2)=$ \{(girl), (girl, young)\}, and $T(S)$ = \{(girl), (table), (girl, young), (girl, sit-at, table)\}. We would like a metric to score $C^2$ higher than $C^1$ because it captures diverse and correct concepts; however, since F1 scores are the same for $c^1_1$, $c^1_2$, $c^2_1$, and is lower for $c^2_2$, the average SPICE of $C^1$ is higher while the oracle SPICE are the same for both $C^1$ and $C^2$.

To address the deficiencies of the existing metrics, we propose a new metric named multi-candidate SPICE (M-SPICE), which takes the \textit{union of the candidate propositions} and computes the F-score against the reference propositions: $F1(T(S), \cup_j T(c_j))$. M-SPICE assigns a higher score if the set captures \textit{diverse and correct} propositions, and it is obvious that the score of $C^2$ is higher than $C^1$ as desired. More comparison between M-SPICE and average SPICE can be found in Section F in the appendix. Figure~\ref{fig:mspice_s2} shows the M-SPICE scores of our SAT-FT model using VQ3 units on SpokenCOCO. When evaluating over multiple captions ($n > 1$), using the beam search hypotheses increases the score less than sampling.

\section{Conclusion}
In this paper, we presented the first model capable of generating fluent spoken captions of images without relying on text, which almost matches the performance of early text-based image captioning models. Our comprehensive experiments demonstrated that learned units need to be robust, of low framerate, and encoding little or none duration information to be a drop-in replacement for text. We also identified the caveats of mode-based evaluation and proposed a new metric to address semantic diversity. As part of this work, a novel dataset of over 600k spoken captions for the MSCOCO dataset is introduced, which we will make publicly available to the research community.

Future work should investigate applying the proposed method to additional languages, devising improved speech unit representations, and jointly training the speech unit model with the I2S model. This would offer the opportunity to explore new analysis-by-synthesis training objectives.

\bibliography{acl2020_main}
\bibliographystyle{acl_natbib}

\appendix

\appendix
\setcounter{table}{0}
\setcounter{figure}{0}
\renewcommand{\thetable}{A\arabic{table}}
\renewcommand{\thefigure}{A\arabic{figure}}

\section{Visually-Grounded Speech Datasets}
\begin{figure*}[htb]
    \centering
    \includegraphics[width=\linewidth]{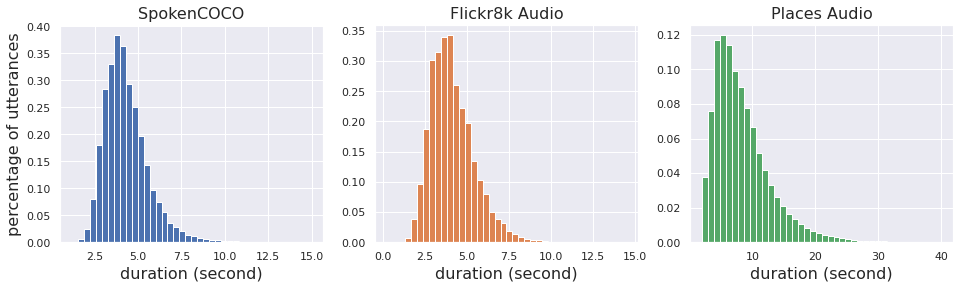}
    \caption{Utterance duration histograms for the three visually-grounded speech datasets.}
    \label{fig:vgs_dur_hist}
\end{figure*}

Table~\ref{tab:datasets} displays details of the three visually-grounded speech datasets used in this paper, and distributions of utterance duration are illustrated in Figure~\ref{fig:vgs_dur_hist}. When computing duration statistics, we exclude utterances longer than 15s for SpokenCOCO and Flickr8k Audio, and 40s for Places Audio, because we found that those utterances resulted from incorrect operation of the data collection interface (e.g., workers forgot to stop recording). When computing vocabulary sizes and word statistics, text transcripts are normalized by lower-casing all the alphabets and removing characters that are neither alphabets nor digits.

For the SpokenCOCO data collection on Amazon Mechanical Turk, we displayed the text of a MSCOCO caption to a user and asked them to record themselves reading the caption out loud. For quality control, we ran a speech recognition system in the background and estimated the word-level transcription for each recording. We computed the word error rate of the ASR output against the text that the user was prompted to read, and only accepted the caption if the word error rate was under 30\%. In the case that the word error rate was higher, the user was asked to re-record their speech. We paid the users \$0.015 per caption recorded, which in conjunction with the 20\% overhead charged by Amazon resulted in a total collection cost of \$10,898.91.
\begin{table*}[htb]
    \centering
    \begin{tabular}{l|cccccc}
    & SpokenCOCO & Flickr8k Audio & Places Audio & \\
    \midrule
    Num. of Utterances & 605495 & 40000 & 400000 \\
    Num. of Speakers & 2353 & 183 & 2683 \\
    Num. of Images & 123287 & 8000 & 400000 \\
    Num. of Utterances / Image & 5 & 5 & 1 \\
    Utterance Duration $\mu$ & 4.12s & 4.33s & 8.37s \\
    Utterance Duration $\sigma$ & 1.31s & 1.33s & 4.53s \\
    Avg. Num. of Words / Utterance & 10.45 & 10.81 & 19.29 \\
    Avg. Num. of Words / Second & 2.41 & 2.63 & 2.31 \\
    Total Duration & 742hr & 46hr & 936hr \\
    Vocabulary Size & 29539 & 8718 & 41217 \\
    Type & scripted & scripted & spontaneous \\
    \end{tabular}
    \caption{\label{tab:datasets}
    Statistics and properties of the three visually-grounded speech datasets used in the paper.}
\end{table*}

\begin{figure*}[htb]
    \centering
    \includegraphics[width=\linewidth]{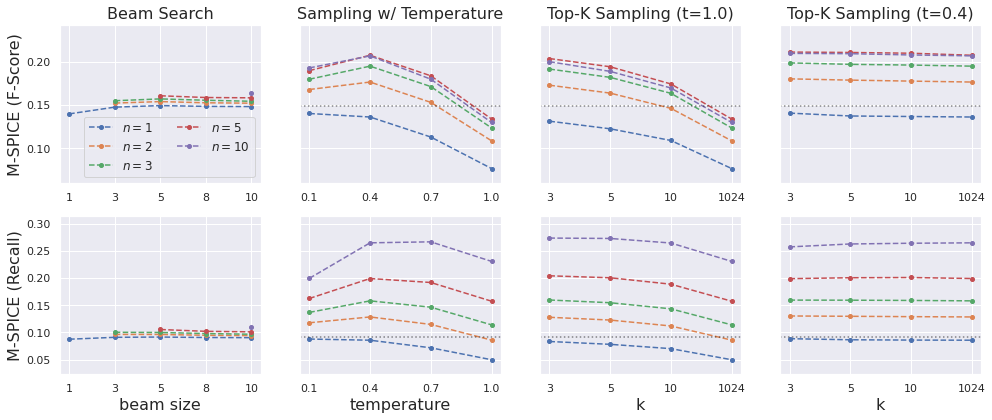}
    \caption{M-SPICE F-score (same as Figure 5) and recall on the SpokenCOCO test set with different candidate proposal methods.}
    \label{fig:mspice_s2_with_rec}
\end{figure*}

\section{Detailed Experimental Setups}
In this section, we provide details about data preprocessing, model architecture, and training hyperparameters for each module used in this paper. The same setups are used for all unit types unless otherwise stated.

\subsection{Image-to-Unit Model}
\paragraph{Data} Images are reshaped to 256$\times$256$\times$3 matrices and per-channel normalized with $\bm{\mu} = [0.485, 0.456, 0.406]$ and $\bm{\sigma}=[0.229, 0.224, 0.225]$. During training, unit sequences are truncated or padded to the target length shown in Table~\ref{tab:i2u_conf}. The target lengths are determined such that there are less than 10\% sequences truncated while still allowing a reasonable batch size to be used. Units that occurred less than five times are excluded. Sequences are not truncated during evaluation. We follow the data splits used in~\citep{harwath_hsu_glass_20} for Places, and \citep{karpathy2015deep} for Flickr8k and SpokenCOCO (commonly known as the ``Karpathy split'').

\begin{table*}[htb]
    \centering
    \begin{tabular}{l|cccccc}
    & Word & Char & VQ3 & VQ2 & WVQ & VQ3 \textbackslash\ RLE \\
    \midrule
Target Length           & 18   & 70   & 100  & 200  & 110  & 160 \\
Sequence Truncated (\%) & 1.12 & 1.74 & 6.90 & 9.37 & 7.80 & 6.35 \\
Batch Size (SAT)        & 80   & 60   & 40   & 40   & 40   & 40 \\
Batch Size (SAT-FT)     & 32   & 32   & 20   & -    & -    & -  \\
    \end{tabular}
    \caption{\label{tab:i2u_conf} Configuration for each type of units used in the Image-to-Unit model.}
\end{table*}

\paragraph{Model} 
We adopt an open-source re-implementation\footnote{\url{https://github.com/sgrvinod/a-PyTorch-Tutorial-to-Image-Captioning}} of Show, Attend, and Tell~\citep{xu2015show} (SAT) with soft attention, which replaces the CNN encoder in~\citep{xu2015show} with a ResNet-101~\citep{he2016deep} pre-trained on ImageNet~\citep{imagenet} for image classification. The last two layers of the ResNet are removed (a pooling layer and a fully-connected layer) such that the encoder produces a $14 \times 14 \times 2048$ feature map for each image.

\paragraph{Training}
Two model variants are considered in this paper: SAT and SAT-FT, which differ in how each part is initialized and which parts are updated during training. The SAT model initializes the encoder parameters with a pre-trained image classification model and freezes the encoder parameters during training. On the other hand, the SAT-FT model (fine-tuned SAT model) initializes the entire model with a pre-trained SAT model, and update all parameters during training. Adam~\citep{kingma2015adam} with a learning rate of $10^{-4}$ is used for optimizing both models. The training objective is maximum likelihood combined with a doubly stochastic attention regularization introduced in~\citep{xu2015show} with a weight of 1. Dropout is applied to the input of decoder softmax layer with a probability of 0.5 during training. Gradients are clipped at 5 for each dimension. The batch size for each unit is shown in Table~\ref{tab:i2u_conf}, which are chosen based on the target length and the GPU memory constraints. All SAT models are trained for at most 30 epochs, and SAT-FT models are trained for at most 20 epochs. Models are selected based on the unit BLEU-4 score on the validation set. 

The time complexity of forward computation is the same for the encoder for all units, while for the decoder it is proportional to the unit sequence length due to the auto-regressive nature. Using two NVIDIA TITAN X Pascal GPUs with data parallel training, each epoch takes about 2.8 hours for VQ3 units and 5.3 hours for VQ2 units.

\subsection{Unit-to-Speech Model}
\paragraph{Data} 
Run-length encoded unit sequences are used as input for all systems (i.e., VQ3 and VQ3 \textbackslash\ RLE systems share the same unit-to-speech model). The native sample rates of audio files in LJSpeech~\citep{ljspeech17} and VCTK~\citep{vctk} are 22050Hz and 48kHz, respectively. For consistency and compatibility with the spectrogram-to-waveform model, we down-sample those in VCTK to 22050Hz. Following Tacotron~\citep{tacotron} and Tacotron2~\citep{tacotron2}, we compute a 80 dimensional Mel spectrogram for each audio file with a 256-sample frame hop, a 1024-sample frame size, and a Hann window function. At a sample rate of 22050Hz, it corresponds to about 11.6ms frame hop and 46.4ms frame size. Utterances longer than 8 seconds are discarded during training to accommodate for the GPU memory constraints. We follow the data splits provided at \url{https://github.com/NVIDIA/tacotron2} for LJSpeech. For the multi-speaker VCTK dataset, to ensure the same speaker distribution between train and valid splits, we randomly sample 2.5\% of the utterances from each speaker for validation, which results in a set of 1087 utterances.

\paragraph{Model}
We use a cascade of two systems to convert a unit sequence into a waveform. The first part synthesizes a Mel spectrogram from a sequence of units, which is referred to as the Unit-to-Speech model in this paper and determines most of the properties of interest in speech (e.g., linguistic content, speaker, prosody). The second part is a vocoder that converts a Mel spectrogram into a waveform, which mostly affects the fidelity of the synthesized speech rather than the aforementioned properties. A vocoder can be either a learned module like WaveNet~\citep{wavenet} and WaveGlow~\citep{waveglow}, or a parameter-less signal processing block such as using the Griffin-Lim algorithm~\citep{griffin1984signal}.

We use an re-implementation\footnote{\url{https://github.com/NVIDIA/tacotron2}} of Tacotron2~\citep{tacotron2} for Unit-to-Speech models, which is a sequence-to-sequence encoder-decoder model with location-sensitive attention~\citep{chorowski2015attention}. For single-speaker models trained on the LJSpeech dataset, the exact same hyperparameters and model architecture are used as~\citep{tacotron2}. For multi-speaker models trained on the VCTK dataset, we create an additional speaker embedding table of 256 dimensions for all speakers and control the speaker identity through these speaker embeddings. Speaker embeddings are injected at two places in the decoder: first in concatenation with the original input to the decoder LSTM, and second in concatenation with the output of the decoder LSTM, right before predicting the stop token and the spectra of a frame.

A pre-trained\footnote{\url{https://github.com/NVIDIA/waveglow}} WaveGlow~\citep{waveglow} vocoder is used for all Unit-to-Speech models, which demonstrates the universality of vocoder models and how little acoustic properties of interest are affected by them. Although it is possible to achieve a even higher fidelity score through training or fine-tuning the WaveGlow model on the re-synthesized spectrograms, we did not attempt to experiment with that, since the focus of this paper is to demonstrate the capability of generating fluent spoken captions and controlling properties like speaker identity independently.

\paragraph{Training}
A batch size of 64 are used for all systems. Adam~\citep{kingma2015adam} with an initial learning rate of $10^{-3}$ is used to minimize the mean square error from spectrogram prediction and the binary cross entropy from stop token prediction combined. L2 regularization for the parameters with a weight of $10^{-6}$ is applied, and the L2 norm of the gradients are clipped at 1. Models are trained for 500 epochs on LJSpeech and 250 epochs on VCTK, and selected based on the validation loss.

The time complexity of forward computation at the encoder is proportional to the unit-sequence length because of the bi-directional LSTM layer in the encoder. On the other hand, the number of the decoding steps is proportional to the duration of the speech at the decoder, which is independent of the choice of input representation; however, at each decoding step, the number of encoder outputs the attention module attends to is proportional to length of the unit sequence. Empirically, each training epoch on LJSpeech takes about 12 minutes using two NVIDIA Titan X Pascal GPUs for both VQ2 and VQ3 models despite that VQ2 sequences are in average twice as long as VQ3 sequences, which shows that time complexity are dominated by other computations.

\begin{table*}[t]
    \centering
    \small
    \begin{tabular}{l|ccc}
& VQ3 & VQ2 & WVQ \\
\midrule
Source Model & ResDAVEnet-VQ & ResDAVEnet-VQ & WaveNet-VQ \\
Training Data & Places Audio+Image & Places Audio+Image & Places Audio \\
Training Objective & Contrastive Loss & Contrastive Loss & Reconstruction Loss \\
RLE ABX Error Rate          & 15.68\% & 13.06\% & 25.23\% \\
Pre-RLE ABX Error Rate      & 14.52\% & 12.51\% & 24.87\% \\
Pre-RLE Unit Rate           & 40ms    & 20ms    & 40ms \\
    \end{tabular}
    \caption{\label{tab:s2u_conf} Properties of the three types of units and the two speech-to-unit models.}
\end{table*}

\begin{table*}[tb]
    \centering
    \small
    \begin{tabular}{l|L{11.5cm}}
        Symbol & Captioned Generated with Beam Search (beam size=5)\\
        \midrule
    VQ3 & 263 32 208 5 336 100 717 803 256 803 815 144 120 144 654 936 48 417 272 417 362 766 825 284 614 156 341 135 769 5 208 32 208 5 336 815 144 815 494 181 467 417 870 395 683 141 250 543 820 587 181 913 1013 467 5 208 32 208 5 467 360 606 360 801 1009 398 847 89 100 869 254 1003 442 42 791 417 272 141 766 362 614 156 341 135 769 5 208 32 \\
    VQ2 & 71 791 71 791 71 791 71 791 71 791 71 791 71 791 71 791 71 791 71 791 71 791 71 791 71 791 71 791 71 791 71 791 71 791...\\
    WVQ & 181 232 181 232 181 232 181 232 181 232 181 232 181 232 181 232 181 232 181 232 181 232 181 232 181 232 181 232 181 232...\\
    VQ3 \textbackslash\ RLE & 263 32 32 32 32 32 32 32 32 32 32 32 32 32 32 32 32 32 32 32 32 32 32 32 32 32 32 32 32 32 32 32 32 32 32 32 32 32 32 32 32...\\
    \end{tabular}
    \caption{Exemplar beam search decoding results from SAT Image-to-Unit models.}
    \label{tab:i2u_beam}
\end{table*}

\begin{table*}[tb]
    \centering
    \small
    \begin{tabular}{l|L{11.5cm}}
        Symbol & Captioned Generated with top-k sampling ($k=5$) \\
        \midrule
VQ3 & 263 208 467 717 288 426 986 72 44 341 151 801 1022 27 320 426 288 66 570 683 351 313 910 820 543 820 230 100 852 329 852 288 502 706 427 110 451 297 938 457 426 100 852 329 852 791 993 522 993 374 502 288 936 48 263 208 32 \\
VQ2 & 71 791 71 791 71 791 71 191 175 51 139 359 173 599 307 419 133 621 85 165 315 883 175 191 71 791 71 48 511 765 983 873 314 409 333 267 409 734 229 787 184 937 886 254 934 666 973 19 947 227 805 967 883 175 48 695 511 655 806 491 647 507 343 867 819 655 699 491 136 221 513 996 675 581 467 652 488 186 3 183 311 613 371 463 314 21 238 910 238 657 230 82 270 868 643 78 391 940 922 49 771 986 147 947 19 957 862 957 95 7 819 695 1011 159 831 589 966 827 753 891 162 253 269 219 13 501 977 302 241 157 691 723 695 175 191 71 791 71 791 71 791 71 48 1007 \\
WVQ & 181 232 181 232 181 232 181 232 181 232 181 225 124 232 181 232 225 232 181 225 124 225 232 181 252 169 211 147 89 67 156 155 189 110 53 246 225 89 52 21 5 216 155 225 25 47 41 223 225 181 166 57 185 82 25 225 124 149 214 93 28 195 65 1 23 109 246 223 141 47 41 223 181 232 82 231 188 169 147 89 225 181 225 124 181 124 5 216 53 246 181 225 137 52 5 159 225 181 225 46 155 246 232 181 232 225 232 181 225 52 30 5 216 166 225 124 225 181 225 5 4 46 225 181 25 137 52 159 155 225 181 225 108 155 246 225 108 155 225 232 181 25 89 221 70 197 232 181 225 214 28 214 225 181 232 244 220 \\
VQ3 \textbackslash\ RLE & 263 32 32 32 32 32 32 32 32 32 32 32 32 32 32 32 208 208 5 5 336 100 803 256 560 417 870 870 870 968 910 250 543 820 587 909 909 181 717 48 936 48 224 176 284 538 133 807 715 39 27 27 476 5 5 476 570 395 395 683 313 141 250 250 587 587 494 909 922 181 100 100 827 119 66 272 417 766 766 766 614 614 156 341 135 135 181 913 913 1013 5 208 208 208 208 208 208 5 5 5 476 320 96 96 651 538 133 766 766 825 740 913 1013 467 5 208 208 32 32 32 32 208 208 5 5 336 501 254 254 254 254 1003 442 852 362 825 740 639 639 587 543 543 975 320 320 284 284 228 844 844 622 622 846 654 654 846 336 263 208 32 32 32 32 32 32 32 32 32 32 32 32 32 32 \\
    \end{tabular}
    \caption{Exemplar sampling results with $(t, k) = (1.0, 5)$ from SAT Image-to-Unit models.}
    \label{tab:i2u_samp}
\end{table*}

\subsection{Speech-to-Unit Model}
We obtain the ResDAVEnet-VQ ``\{2\} $\rightarrow$ \{2, 3\}'' model and the WaveNet-VQ (PA) model reported in~\cite{harwath_hsu_glass_20} from the authors. Both models learn discrete representations for speech and are used to transcribe speech into a sequence of units in this paper. We use these models to extract unit sequences for all datasets without fine-tuning, which examines the robustness of these Speech-to-Unit models when applied to datasets of different domains. Table~\ref{tab:s2u_conf} compares the three types of units used in this paper (VQ3, VQ2, WVQ) extracted from these two models. For self-containedness, the ABX error rate for each unit reported in~\citep{harwath_hsu_glass_20} are also included. The ABX test evaluates the phone discriminability of the learned units on the ZeroSpeech 2019 English test set~\citep{zerospeech19}. Note that (1) VQ3 and WVQ have the same unit rate before run-length encoding, (2) VQ2 achieves the lowest ABX error rate, (3) and all units have a lower ABX error rate before run-length encoding.

% \clearpage
\section{Image-to-Unit Samples}
Table~\ref{tab:i2u_beam} and \ref{tab:i2u_samp} display unit captions for the same image generated from Image-to-Unit models trained on different learned units. Captions in Table~\ref{tab:i2u_beam} are decoded with beam search (beam size=5), and those in Table~\ref{tab:i2u_samp} are sampled from the model distribution with top-k sampling ($k=5$). 

Table~\ref{tab:i2u_beam} shows that Image-to-Unit models trained on WVQ and VQ3 units without run-length encoding (VQ3 \textbackslash\ RLE) fail to produce reasonable captions using beam search for \emph{all} images. In fact, generated captions are almost always the same among different images for these two models. The WVQ model generates captions looping the same bi-gram until exceeding the maximum length, while the VQ3 \textbackslash\ RLE repeats the same unit. On the other hand, the model trained on VQ2 units can sometimes produce reasonable captions, but it exhibit the same behavior as the WVQ model when it fails as shown here.

\begin{figure*}[htb!]
    \centering
    \includegraphics[width=\linewidth]{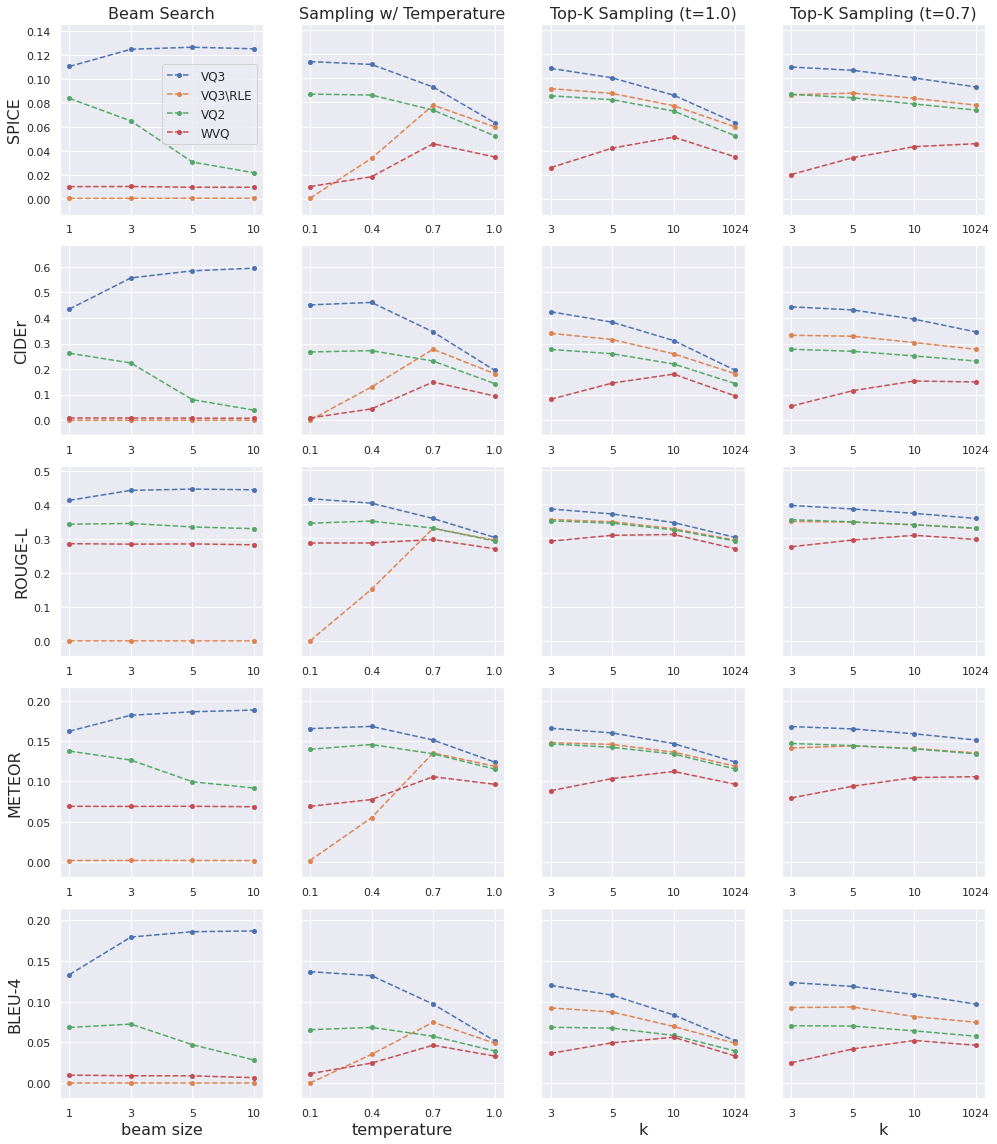}
    \caption{Word-based caption evaluation results of all five metrics on SAT models trained on different units, using both beam search decoding and sampling for unit sequence generation.}
    \label{fig:unit_full}
\end{figure*}

On the contrary, Table~\ref{tab:i2u_samp} shows that all four models are capable of generating non-trivial captions without looping via sampling.\footnote{The model might still generate trivial captions when sampling with a temperature close to zero or setting a very small $k$ with top-k sampling, in which case sampling is similar to greedy decoding.} The observation here is consistent with the evaluation results on the transcribed spoken captions presented in Table~\ref{tab:cap_eval_beam_all_sym} and \ref{tab:cap_eval_samp_unit} and Figure~\ref{fig:unit_full}.

\begin{table*}[htb]
    \centering
    % \resizebox{.95\linewidth}{!}{
    \small
    \begin{tabular}{c|cccc}
        \toprule
        \multirow{2}{*}{symbol} 
        & \multicolumn{4}{c}{Greedy / Beam-Search (SAT Model)} \\
        & Unit BLEU-4 & Unit METEOR & Unit ROUGE & Unit CIDER \\
        \midrule\midrule
VQ3                     & 0.176 / 0.274 & 0.178 / 0.196 & 0.280 / 0.328 & 0.121 / 0.215 \\
VQ2                     & 0.172 / 0.141 & 0.132 / 0.108 & 0.178 / 0.157 & 0.027 / 0.020 \\
WVQ                     & 0.019 / 0.020 & 0.048 / 0.048 & 0.081 / 0.081 & 0.000 / 0.000 \\
VQ3 \textbackslash\ RLE & 0.163 / 0.163 & 0.168 / 0.168 & 0.218 / 0.218 & 0.000 / 0.000 \\
        \bottomrule
    \end{tabular}
    \caption{\label{tab:cap_unit_eval_beam_all_sym} Unit-based caption evaluation on MSCOCO test set. The beam size $\in$ \{3, 5, 10\} was chosen for each model to maximize the CIDEr score. Note that the scores between different units are not directly comparable, because they are computed based different types of units.}
\end{table*}

% \clearpage
\section{Caption Evaluation on Unit Sequences}
The spoken caption evaluations presented in Section 3.2 utilized an automatic speech recognizer to transcribe the generated captions into words so that they could be compared against reference text captions. In the case that a speech recognizer is not available, we can perform evaluation directly on the speech unit representations by using the unit model to transcribe the reference spoken captions. Table~\ref{tab:cap_unit_eval_beam_all_sym} displays BLEU-4, METEOR, ROUGE, and CIDEr scores evaluated directly on the various speech unit representations; we cannot compute SPICE scores directly on the speech units because SPICE requires a dependency tree for each caption. It is important to note that for a given evaluation metric, the scores across the models are not directly comparable because their unit spaces are different. We do note that the relative ranking among VQ3, VQ2, and Wavenet-VQ is consistent across BLEU-4, METEOR, and ROUGE, however, VQ3 \textbackslash\ RLE achieves abnormally high scores on these metrics despite producing trivial captions for all images as shown in Table~\ref{tab:i2u_beam}. This is because unit ``32'' has learned to represent non-speech frames such as silence, which frequently occurs at both the beginning and end of utterances. Without RLE, consecutive strings of ``32'' units are extremely common in both the candidate and reference captions, which inflates the scores of this model. The exception here is the CIDEr metric, which incorporates TF-IDF weighting that tends to de-emphasize these kinds of uninformative patterns. The fact that the CIDEr score is 0 for both the VQ3 \textbackslash\ RLE and Wavenet-VQ models indicates that in general the captions produced by these models are uninformative. We posit that word-level evaluation is always preferable for spoken caption generation, but in the case that this is not possible the CIDEr metric may be the best option.

\begin{table*}[htb]
    \centering
    \small
    \begin{tabular}{c|ccccc}
        \toprule
        \multirow{2}{*}{symbol} 
        & \multicolumn{5}{c}{Greedy / Beam-Search (SAT Model)} \\
        & BLEU-4 & METEOR & ROUGE & CIDEr & SPICE \\
        \midrule\midrule
word     & 0.287 / 0.315 & 0.247 / 0.253 & 0.524 / 0.533 & 0.939 / 0.984 & 0.180 / 0.185 \\
char     & 0.238 / 0.289 & 0.230 / 0.239 & 0.495 / 0.512 & 0.783 / 0.879 & 0.164 / 0.172 \\
VQ3      & 0.133 / 0.186 & 0.162 / 0.186 & 0.413 / 0.446 & 0.435 / 0.584 & 0.111 / 0.127 \\
VQ2      & 0.068 / 0.073 & 0.138 / 0.126 & 0.343 / 0.345 & 0.262 / 0.224 & 0.084 / 0.065 \\
WVQ      & 0.010 / 0.009 & 0.069 / 0.069 & 0.286 / 0.285 & 0.009 / 0.009 & 0.011 / 0.011 \\
VQ3 \textbackslash\ RLE & 0.000 / 0.000 & 0.002 / 0.002 & 0.001 / 0.001 & 0.000 / 0.000 & 0.001 / 0.001 \\
        \bottomrule
    \end{tabular}
    \caption{Word-based caption evaluation on MSCOCO test set. An ASR model is used to transcribe the spoken captions into text for evaluation. The beam size $\in \{3, 5, 10\}$ was chosen for each model to maximize the SPICE score.}
    \label{tab:cap_eval_beam_all_sym}
\end{table*}
\begin{table*}[htb]
    \centering
    \resizebox{\linewidth}{!}{
    \small
    \begin{tabular}{cc|cccc|ccc|ccc}
        \toprule
\multirow{2}{*}{Metric} & \multirow{2}{*}{symbol}
& \multicolumn{4}{c}{Sampling with Temperature} 
& \multicolumn{3}{|c}{Top-K Sampling ($t=1.0$)} 
& \multicolumn{3}{|c}{Top-K Sampling ($t=0.7$)} \\
& & $t=1.0$ & $t=0.7$ & $t=0.4$ & $t=0.1$ & $k=10$ & $k=5$ & $k=3$ & $k=10$ & $k=5$ & $k=3$ \\
        \midrule\midrule
\multirow{4}{*}{BLEU-4}
& VQ3                    & 0.052 & 0.097 & 0.132 & 0.137 & 0.084 & 0.108 & 0.120 & 0.109 & 0.119 & 0.124 \\
& VQ2                    & 0.039 & 0.058 & 0.068 & 0.066 & 0.059 & 0.068 & 0.069 & 0.064 & 0.070 & 0.071 \\
& WVQ                    & 0.033 & 0.047 & 0.025 & 0.012 & 0.056 & 0.050 & 0.037 & 0.052 & 0.042 & 0.025 \\
& VQ3 \textbackslash\ RLE & 0.049 & 0.075 & 0.035 & 0.000 & 0.070 & 0.087 & 0.092 & 0.082 & 0.094 & 0.093 \\
        \midrule\midrule
\multirow{4}{*}{METEOR}
& VQ3                    & 0.124 & 0.151 & 0.168 & 0.165 & 0.147 & 0.160 & 0.166 & 0.159 & 0.165 & 0.168\\
& VQ2                    & 0.115 & 0.134 & 0.146 & 0.140 & 0.134 & 0.142 & 0.147 & 0.140 & 0.144 & 0.147\\
& WVQ                    & 0.096 & 0.106 & 0.078 & 0.069 & 0.112 & 0.104 & 0.088 & 0.105 & 0.094 & 0.080\\
& VQ3 \textbackslash\ RLE & 0.119 & 0.135 & 0.055 & 0.002 & 0.136 & 0.146 & 0.148 & 0.141 & 0.144 & 0.141\\
        \midrule\midrule
\multirow{4}{*}{ROUGE-L}
& VQ3                    & 0.303 & 0.358 & 0.403 & 0.416 & 0.346 & 0.371 & 0.386 & 0.373 & 0.386 & 0.397\\
& VQ2                    & 0.293 & 0.330 & 0.351 & 0.345 & 0.325 & 0.345 & 0.351 & 0.340 & 0.348 & 0.355\\
& WVQ                    & 0.270 & 0.297 & 0.287 & 0.287 & 0.312 & 0.309 & 0.292 & 0.309 & 0.295 & 0.276\\
& VQ3 \textbackslash\ RLE & 0.295 & 0.330 & 0.152 & 0.001 & 0.328 & 0.349 & 0.355 & 0.340 & 0.348 & 0.350\\
        \midrule\midrule
\multirow{4}{*}{CIDEr}
& VQ3                    & 0.195 & 0.345 & 0.461 & 0.451 & 0.312 & 0.383 & 0.424 & 0.395 & 0.431 & 0.444\\
& VQ2                    & 0.143 & 0.231 & 0.272 & 0.267 & 0.220 & 0.260 & 0.277 & 0.251 & 0.270 & 0.278\\
& WVQ                    & 0.095 & 0.150 & 0.044 & 0.009 & 0.180 & 0.145 & 0.082 & 0.154 & 0.116 & 0.055\\
& VQ3 \textbackslash\ RLE & 0.182 & 0.277 & 0.130 & 0.000 & 0.260 & 0.316 & 0.340 & 0.304 & 0.328 & 0.332\\
        \midrule\midrule
\multirow{4}{*}{SPICE}
& VQ3                    & 0.063 & 0.093 & 0.111 & 0.114 & 0.086 & 0.100 & 0.108 & 0.100 & 0.106 & 0.109\\
& VQ2                    & 0.052 & 0.074 & 0.086 & 0.087 & 0.073 & 0.082 & 0.085 & 0.079 & 0.084 & 0.087\\
& WVQ                    & 0.035 & 0.046 & 0.019 & 0.011 & 0.051 & 0.042 & 0.026 & 0.043 & 0.034 & 0.020\\
& VQ3 \textbackslash\ RLE & 0.060 & 0.078 & 0.034 & 0.001 & 0.077 & 0.087 & 0.091 & 0.083 & 0.088 & 0.086\\
        \bottomrule
    \end{tabular}
    }
    \caption{Sampling-based evaluations with SAT models trained on different units.}
    \label{tab:cap_eval_samp_unit}
\end{table*}

\section{Full Results of Caption Evaluation on Word Sequences}
Table~\ref{tab:cap_eval_beam_all_sym} and \ref{tab:cap_eval_samp_unit} and Figure~\ref{fig:unit_full} present the complete caption evaluation results on transcribed word sequences for all 5 metrics, supplementing Figure 3 in the main paper that only presents the SPICE results. Note that the transcripts for WVQ and VQ3 \textbackslash\ RLE beam search captions are not reliable; for the reasons discussed in the previous section the generated spoken captions contain only silence, and the ASR model used for transcription did not see utterances comprised of pure silence during training. We see that ranking between symbols are generally consistent among all those metrics, except the ranking between WVQ and VQ3 \textbackslash\ RLE when sampling with a temperature of 0.4. This is a relatively low-score regime when both model are transiting from generating trivial caption ($t=0.1$) to non-trivial captions ($t=0.7$).

% \clearpage
\section{Comparison of Average SPICE and M-SPICE}
\begin{figure*}[htb]
    \centering
    \includegraphics[width=\linewidth]{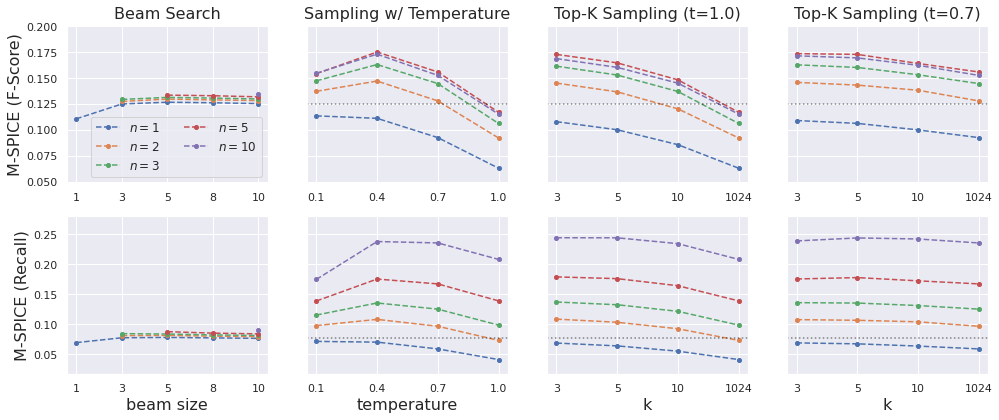}
    \includegraphics[width=.8\linewidth]{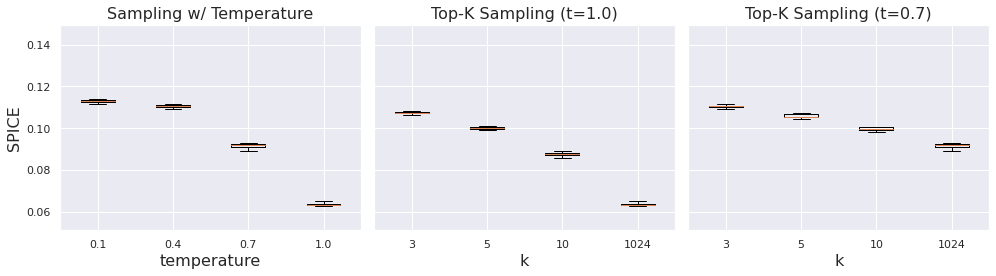}
    \caption{M-SPICE F1-scores and recalls (top) and SPICE distributions (bottom) of the SAT model on the MSCOCO test set with different caption generation methods. Box-and-whisker plots the SPICE scores over 10 runs are shown, where a box extends from the first quartile to the third quartile. }
    \label{fig:avg_spice_mspice_sat}
\end{figure*}

\begin{figure*}[htb]
    \centering
    \includegraphics[width=\linewidth]{figs/mspice_stage2_lg.png}
    \includegraphics[width=.8\linewidth]{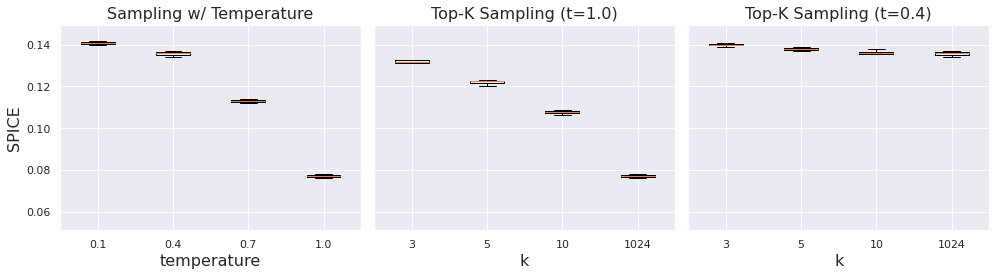}
    \caption{M-SPICE F1-scores and recalls (top) and SPICE distributions (bottom) of the SAT-FT model on the MSCOCO test set with different caption generation methods.}
    \label{fig:avg_spice_mspice_satft}
\end{figure*}

Figures~\ref{fig:avg_spice_mspice_sat} and \ref{fig:avg_spice_mspice_satft} display the M-SPICE scores and SPICE score distributions of different sampling methods for the SAT and SAT-FT model trained on VQ3 units, respectively. The exact numbers are shown in Tables~\ref{tab:mspice_sat_beam}, \ref{tab:mspice_sat_samp}, \ref{tab:mspice_satft_beam}, and \ref{tab:mspice_satft_samp}. When performing sampling-based evaluation, there is bound to be some stochasticity in the results. However, the SPICE score distributions (box plots over 10 trials) shown in the bottom row of Figures~\ref{fig:avg_spice_mspice_sat} and \ref{fig:avg_spice_mspice_satft} are very narrow, which we attribute to the fact that the COCO test set is large enough attenuate this stochasticity. The narrowness of the box plots also suggests that taking the average SPICE score over multiple sampling runs does not reflect the diversity of the captions the way that M-SPICE does.

\begin{table*}[htb]
    \begin{minipage}{0.48\linewidth}
        \centering
        \small
        \begin{tabular}{c|ccccc}
            \toprule
    \multirow{2}{*}{$n=$}
    & \multicolumn{5}{c}{Beam Size} \\
    & 1 & 3 & 5 & 8 & 10 \\
            \midrule\midrule
    1   & 0.111 & 0.125 & 0.127 & 0.126 & 0.125 \\
    2   & -     & 0.127 & 0.130 & 0.129 & 0.128 \\
    3   & -     & 0.129 & 0.131 & 0.131 & 0.130 \\
    5   & -     & -     & 0.134 & 0.133 & 0.132 \\
    10  & -     & -     & -     & -     & 0.135 \\
            \bottomrule
        \end{tabular}
        \caption{M-SPICE F1-scores of the VQ3 SAT model with beam search decoding.}
        \label{tab:mspice_sat_beam}
    \end{minipage}
    \hfill
    \begin{minipage}{0.48\linewidth}
        \centering
        \small
        \begin{tabular}{c|ccccc}
            \toprule
    \multirow{2}{*}{$n=$}
    & \multicolumn{5}{c}{Beam Size} \\
    & 1 & 3 & 5 & 8 & 10 \\
            \midrule\midrule
    1   & 0.140 & 0.147 & 0.149 & 0.148 & 0.148 \\
    2   & -     & 0.152 & 0.154 & 0.152 & 0.152 \\
    3   & -     & 0.155 & 0.157 & 0.155 & 0.154 \\
    5   & -     & -     & 0.161 & 0.159 & 0.158 \\
    10  & -     & -     & -     & -     & 0.164 \\
            \bottomrule
        \end{tabular}
        \caption{M-SPICE F1-scores of the VQ3 SAT-FT model with beam search decoding.}
        \label{tab:mspice_satft_beam}
    \end{minipage}
    
    \vspace{.4cm}
    
    \centering
    \resizebox{\linewidth}{!}{
    \small
    \begin{tabular}{c|cccc|ccc|ccc}
        \toprule
\multirow{2}{*}{$n=$}
& \multicolumn{4}{c}{Sampling with Temperature} 
& \multicolumn{3}{|c}{Top-K Sampling ($t=1.0$)} 
& \multicolumn{3}{|c}{Top-K Sampling ($t=0.4$)} \\
& $t=1.0$ & $t=0.7$ & $t=0.4$ & $t=0.1$ & $k=10$ & $k=5$ & $k=3$ & $k=10$ & $k=5$ & $k=3$ \\
        \midrule\midrule
1   & 0.063 & 0.093 & 0.111 & 0.114 & 0.086 & 0.100 & 0.108 & 0.100 & 0.106 & 0.109 \\
2   & 0.092 & 0.128 & 0.147 & 0.137 & 0.120 & 0.137 & 0.145 & 0.138 & 0.143 & 0.146 \\
3   & 0.106 & 0.145 & 0.163 & 0.147 & 0.137 & 0.153 & 0.161 & 0.153 & 0.160 & 0.163 \\
5   & 0.117 & 0.156 & 0.175 & 0.154 & 0.148 & 0.165 & 0.173 & 0.164 & 0.173 & 0.174 \\
10  & 0.115 & 0.153 & 0.173 & 0.155 & 0.145 & 0.160 & 0.169 & 0.162 & 0.169 & 0.171 \\
        \bottomrule
    \end{tabular}
    }
    \caption{M-SPICE F1-scores of the VQ3 SAT model with sampling. $t$ denotes the temperature, and $k$ denotes the number of top units considered at each decoding step for top-K sampling.}
    \label{tab:mspice_sat_samp}
    
    \vspace{.3cm}
    
    \centering
    \resizebox{\linewidth}{!}{
    \small
    \begin{tabular}{c|cccc|ccc|ccc}
        \toprule
\multirow{2}{*}{$n=$}
& \multicolumn{4}{c}{Sampling with Temperature} 
& \multicolumn{3}{|c}{Top-K Sampling ($t=1.0$)} 
& \multicolumn{3}{|c}{Top-K Sampling ($t=0.4$)} \\
& $t=1.0$ & $t=0.7$ & $t=0.4$ & $t=0.1$ & $k=10$ & $k=5$ & $k=3$ & $k=10$ & $k=5$ & $k=3$ \\
        \midrule\midrule
1   & 0.076 & 0.113 & 0.136 & 0.140 & 0.109 & 0.122 & 0.131 & 0.137 & 0.137 & 0.141 \\
2   & 0.108 & 0.153 & 0.177 & 0.168 & 0.146 & 0.164 & 0.173 & 0.178 & 0.179 & 0.180 \\
3   & 0.123 & 0.171 & 0.195 & 0.180 & 0.163 & 0.182 & 0.192 & 0.196 & 0.197 & 0.199 \\
5   & 0.134 & 0.184 & 0.208 & 0.190 & 0.174 & 0.194 & 0.204 & 0.210 & 0.211 & 0.211 \\
10  & 0.130 & 0.180 & 0.207 & 0.193 & 0.170 & 0.189 & 0.200 & 0.208 & 0.209 & 0.210 \\
        \bottomrule
    \end{tabular}
    }
    \caption{M-SPICE F1-scores of the VQ3 SAT-FT model with sampling.}
    \label{tab:mspice_satft_samp}
\end{table*}

% \vspace{.5cm}

\begin{table*}[htb!]
    \centering
    \small
    \resizebox{\linewidth}{!}{
    \begin{tabular}{c|ccccc|cccc|ccc|ccc}
        \toprule
\multirow{3}{*}{$n$}
& \multicolumn{5}{c|}{Beam Search}
& \multicolumn{10}{c}{Sampling ($t$: temperature; $k$: top-k)} \\
& \multicolumn{5}{c|}{beam size=?}
& \multicolumn{4}{c|}{$(t, k)=(?, All)$}
& \multicolumn{3}{c|}{$(t, k)=(1.0, ?)$}
& \multicolumn{3}{c}{$(t, k)=(0.7, ?)$} \\
& 1 & 3 & 5 & 8 & 10 
& $1.0$ & $0.7$ & $0.4$ & $0.1$ 
& $10$ & $5$ & $3$ & $10$ & $5$ & $3$ \\
        \midrule\midrule
1  & 551 & 479 & 447 & 421 & 411 & 1447 &  978 &  689 &  561  & 1058 &  908 &  770 &  694 &  663 & 670 \\
2  & -   & 572 & 523 & 502 & 474 & 2100 & 1367 &  917 &  696  & 1522 & 1289 & 1025 &  907 &  867 & 851 \\
3  & -   & 693 & 620 & 585 & 562 & 2550 & 1644 & 1075 &  803  & 1855 & 1515 & 1222 & 1069 & 1003 & 973 \\
5  & -   & -   & 681 & 625 & 617 & 3239 & 2111 & 1305 &  938  & 2367 & 1861 & 1511 & 1266 & 1209 & 1155 \\
10 & -   & -   & -   & -   & 700 & 4311 & 2876 & 1664 & 1155  & 3176 & 2512 & 1954 & 1618 & 1552 & 1437 \\
        \bottomrule
    \end{tabular}
    }
    \caption{The vocabulary size of the VQ3 SAT-FT model as estimated by various decoding approaches. The numbers in this table display the specific values of the curves depicted in Figure 4.}
    \label{tab:vocab_beam}
\end{table*}
\section{Full Results of Learned Vocabulary Size}
In Table~\ref{tab:vocab_beam}, we display the numerical results depicted graphically in Figure 4.

% \clearpage
\begin{table*}[htb]
\small
    \centering
    \resizebox{.95\linewidth}{!}{
    \begin{tabular}{cccc|ccccc}
        \toprule
        Train Data & Speaker ID & Gender & Region &
        BLEU-4 & METEOR & ROUGE & CIDER & SPICE \\
        \midrule\midrule
LJSpeech & - & F & -  & 0.233 & 0.212 & 0.478 & 0.732 & 0.149 \\
        \midrule
\multirow{5}{*}{VCTK}
& p247 & M & Scottish & 0.234 & 0.211 & 0.480 & 0.730 & 0.148 \\
& p231 & F & English  & 0.233 & 0.210 & 0.478 & 0.724 & 0.146 \\
& p294 & F & American & 0.236 & 0.212 & 0.482 & 0.732 & 0.148 \\
& p345 & M & American & 0.234 & 0.209 & 0.477 & 0.717 & 0.144 \\
& p307 & F & Canadian & 0.234 & 0.211 & 0.479 & 0.729 & 0.148 \\
        \bottomrule
    \end{tabular}
    }
    \caption{Demonstration of disentangled voice control via synthesizing the same units with different unit-to-speech models conditioned on different speaker IDs. Units are generated with beam search using the SAT-FT model for the MSCOCO test set.}
    \label{tab:voice_control}
\end{table*}
\section{Disentangled Voice Control for Image-to-Speech Synthesis}
We examine to what extent the VQ3 units are portable across different speakers by training a U2S model on the VCTK dataset that additionally takes a speaker ID as input.
%, using speech from 5 speakers from VCTK dataset. 
The resulting model is able to generate speech with the voice of any VCTK speaker. We evaluate the captions produced by this system on SpokenCOCO for 5 speakers in Table~\ref{tab:voice_control}. In order to compute these scores we transcribe the captions generated by each model into text using the ASR system we describe in Section~\ref{sec:experiments}, which was solely trained on re-synthesized SpokenCOCO captions using the LJSpeech U2S model. The scores in Table~\ref{tab:voice_control} indicate not only that the I2U model can be easily integrated with U2S models representing a diverse set of speakers, but also that the LJSpeech ASR system works very well on the speech synthesized from the VCTK models. In Figure~\ref{fig:example}, we show example captions generated by conditioning on the same unit sequence, but different speaker identities.

% \clearpage
\section{More Image-to-Speech Samples}
In Tables~\ref{tab:i2s_demo_beam} and \ref{tab:i2s_demo_samp}, we show many more examples of spoken captions generated by the VQ3 model. In Table~\ref{tab:i2s_demo_beam}, all three captions in each row were generated from the same unit sequence corresponding to the top hypothesis from beam search decoding. Each column represents the caption waveform generated by a different U2S model reflecting a different speaker. Although the spectrograms are visibly very different (reflecting the differing speaker characteristics across the U2S models), the word sequence estimated by the ASR model is generally the same. We do notice some substitution errors (highlighted in red), most commonly between ``a'' and ``the''.

Table~\ref{tab:i2s_demo_samp} shows captions generated by the same VQ3 model and for the same set of images depicted in Table~\ref{tab:i2s_demo_beam}, but instead of varying the U2S model we show captions generated via sampling rather than beam search. Here, we note that the sampled captions exhibit diversity both their content and linguistic style. We observe that the captioning model has learned to produce captions that correctly use quantifiers and conjugate verbs (``a couple of cows walking'' vs. ``a cow is standing''). The model also disentangles object identity from attributes such as color ``red fire hydrant'' vs. ``yellow fire hydrant'' vs. ``green fire hydrant'').
\begin{table*}[htb]
    \centering
    \small
    \begin{tabular}{C{2cm}|C{3.3cm}|C{3.3cm}|C{3.3cm}}
    \multirow{2}{*}{Image}
    & \multicolumn{3}{c}{Generated Spoken Captions / Transcripts (SAT-FT, VQ3, Beam Search)} \\
    & LJSpeech & VCTK (p247) & VCTK (p307) \\
    \midrule\midrule
    \raisebox{-.65\height}{\includegraphics[width=\linewidth]{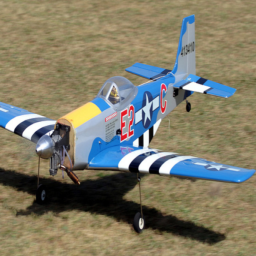}}
    & \plotspec{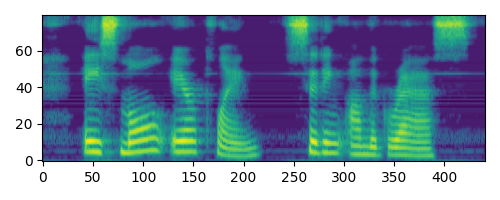} a small airplane sitting on the grass
    & \plotspec{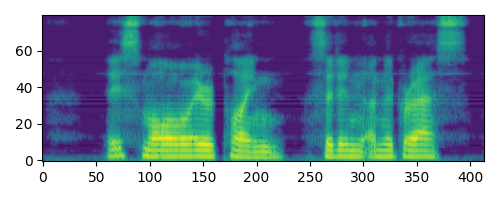} a small airplane sitting on the grass
    & \plotspec{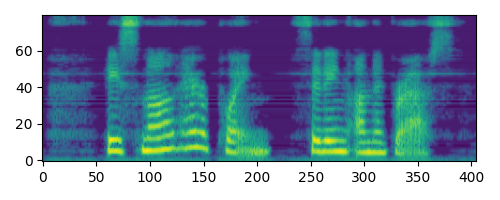} a small airplane sitting on the grass \\
    \midrule
    \raisebox{-.65\height}{\includegraphics[width=\linewidth]{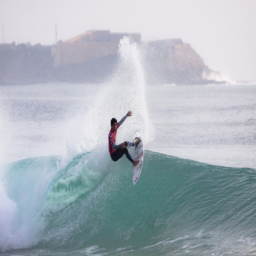}}
    & \plotspec{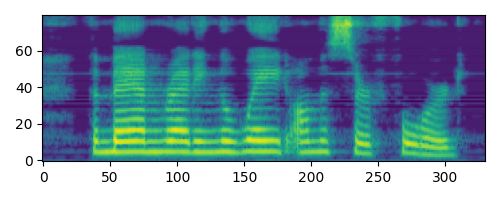} a man riding a wave on a surfboard
    & \plotspec{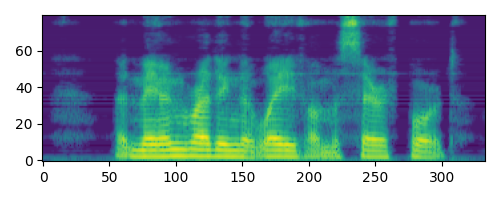} a man riding a wave on a surfboard
    & \plotspec{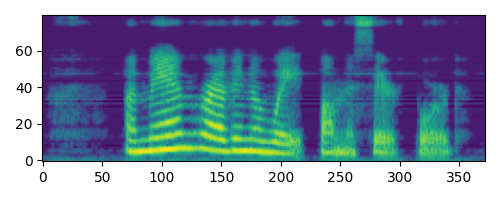} a man riding a wave on a surfboard \\
    \midrule
    \raisebox{-.65\height}{\includegraphics[width=\linewidth]{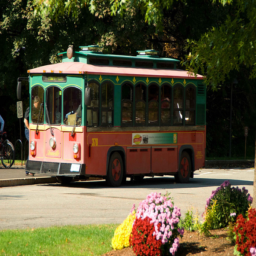}}
    & \plotspec{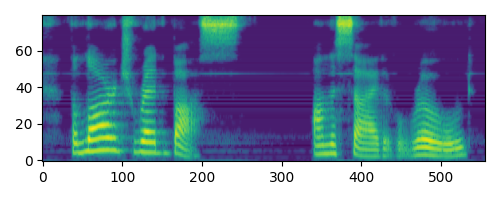} a large red bus on the side of \red{the} road
    & \plotspec{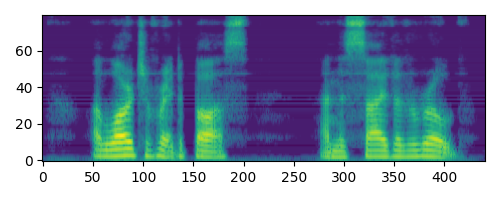} a large red bus on the side of \red{the} road
    & \plotspec{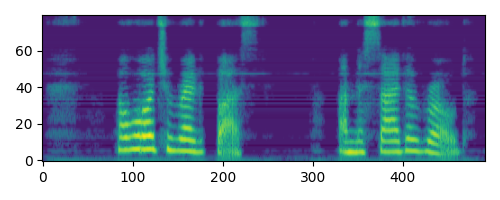} a large red bus on the side of \red{a} road \\
    \midrule
    \raisebox{-.65\height}{\includegraphics[width=\linewidth]{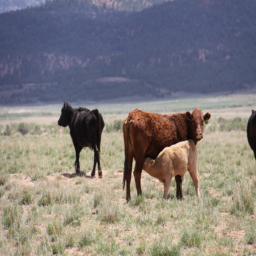}}
    & \plotspec{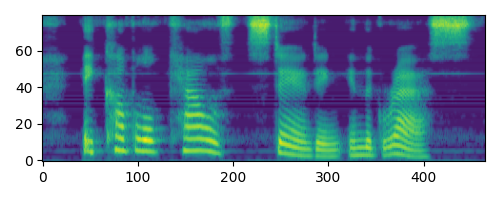} a couple of cows standing in the grass
    & \plotspec{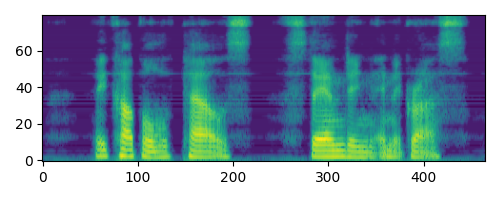} a couple of cows standing in the grass
    & \plotspec{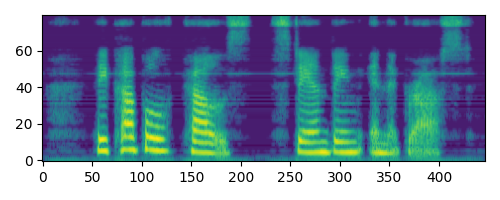} a couple of cows standing in the grass \\
    \midrule
    \raisebox{-.65\height}{\includegraphics[width=\linewidth]{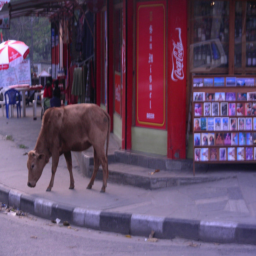}}
    & \plotspec{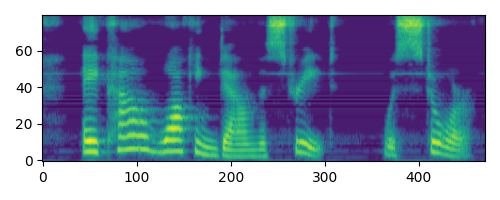} a cow walking down the street \red{with} a store
    & \plotspec{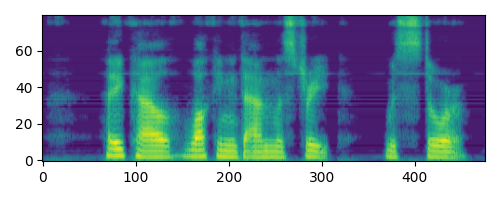} a cow walking down the street \red{in} a store
    & \plotspec{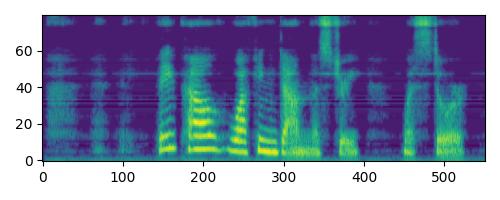} a cow walking down the street \red{next to} a store \\
    \midrule
    \raisebox{-.65\height}{\includegraphics[width=\linewidth]{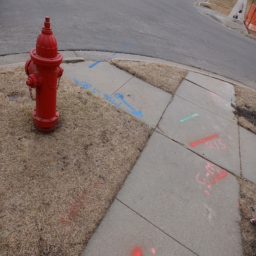}}
    & \plotspec{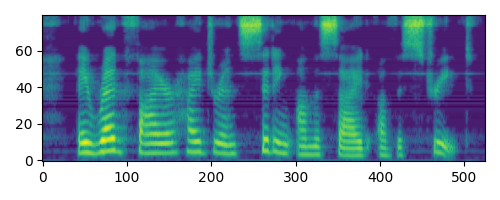} a red fire hydrant sitting on the side of a street
    & \plotspec{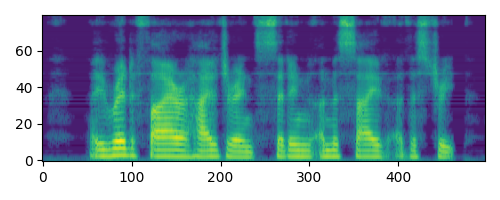} a red fire hydrant sitting on the side of a street
    & \plotspec{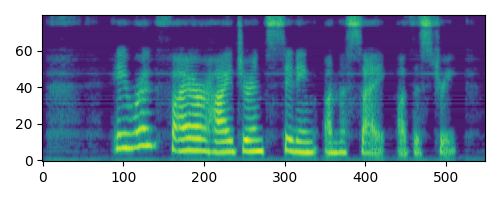} a red fire hydrant sitting on the side of a street\\
    \midrule
    \raisebox{-.65\height}{\includegraphics[width=\linewidth]{sample/37069.png}}
    & \plotspec{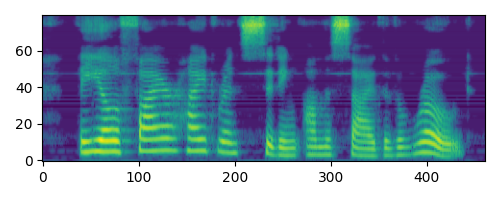} a yellow fire hydrant sitting on the side of \red{a} road
    & \plotspec{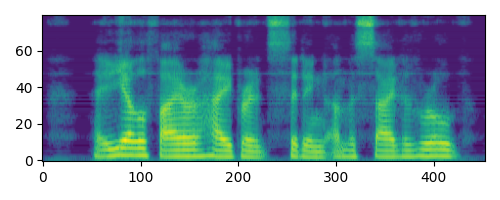} a yellow fire hydrant sitting on the side of \red{the} road
    & \plotspec{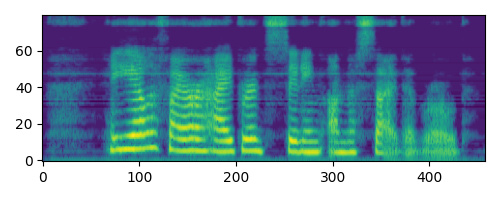} a yellow fire hydrant sitting on the side of \red{a} road \\
    \midrule
    \raisebox{-.65\height}{\includegraphics[width=\linewidth]{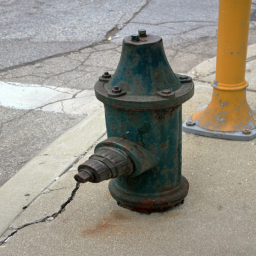}}
    & \plotspec{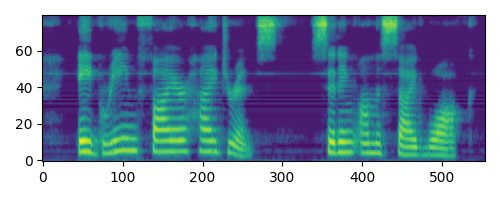} a green fire hydrant sitting on \red{a} sidewalk
    & \plotspec{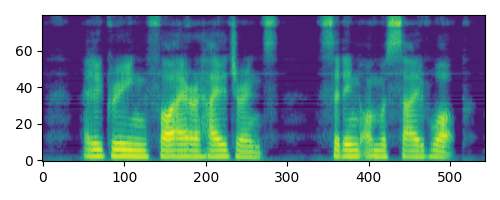} a green fire hydrant sitting on \red{the} sidewalk
    & \plotspec{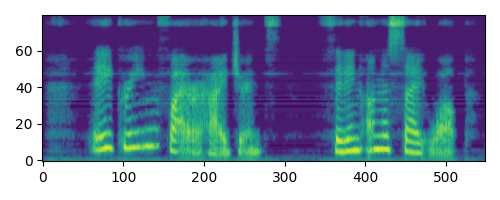} a green fire hydrant sitting on \red{a} sidewalk \\
    \end{tabular}
    \caption{Samples. More at \url{https://wnhsu.github.io/image-to-speech-demo/3_vq3_voice_control_sat-ft_model}}
    \label{tab:i2s_demo_beam}
\end{table*}

\begin{table*}[htb]
    \centering
    \small
    \begin{tabular}{C{2cm}|C{3.3cm}|C{3.3cm}|C{3.3cm}}
    \multirow{2}{*}{Image}
    & \multicolumn{3}{c}{Generated Spoken Captions / Transcripts (SAT-FT, VQ3, Sampling $(t, k)=(0.4, 3)$)} \\
    & trial 1 & trial 2 & trial 3 \\
    \midrule\midrule
    \raisebox{-.65\height}{\includegraphics[width=\linewidth]{sample/155.png}}
    & \plotspec{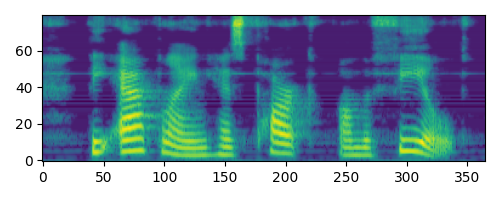} the airplane is parked on the field
    & \plotspec{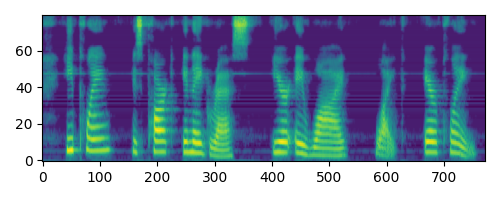} a plane is parked in the grass near a white and white airplane
    & \plotspec{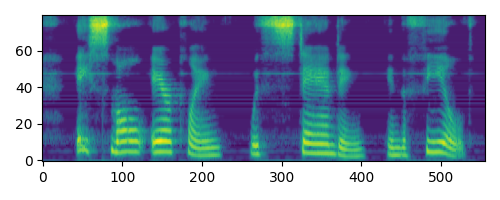} a small airplane that is standing in a field \\
    \midrule
    \raisebox{-.65\height}{\includegraphics[width=\linewidth]{sample/15061.png}}
    & \plotspec{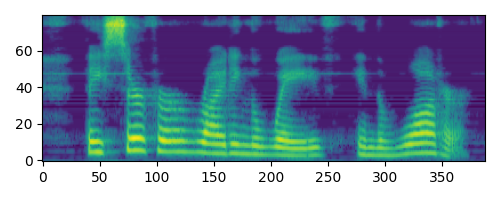} a surfer riding a wave in the water
    & \plotspec{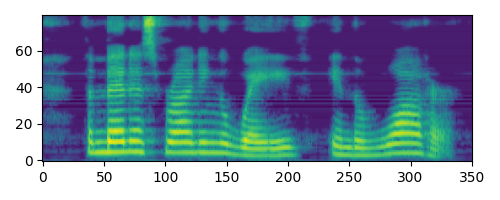} the man is riding the wave in the water
    & \plotspec{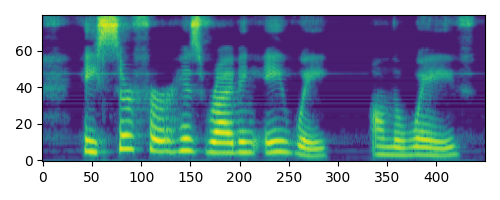} a surfer is riding a wave on a wave \\
    \midrule
    \raisebox{-.65\height}{\includegraphics[width=\linewidth]{sample/24752.png}}
    & \plotspec{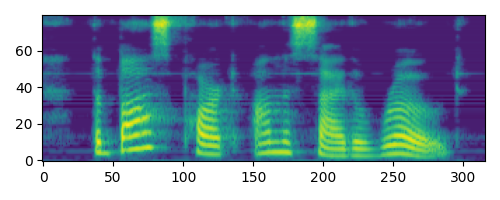} the bus parked on the side of the road
    & \plotspec{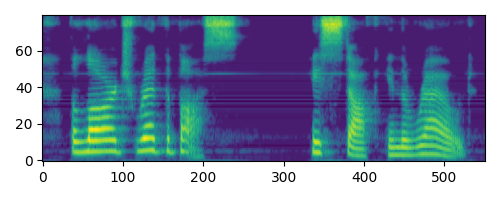} a large red bus is stopped in the road
    & \plotspec{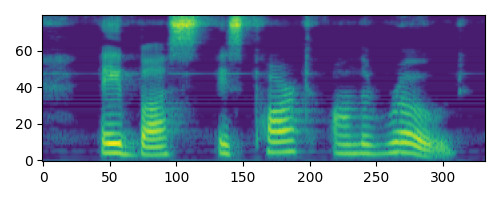} a bus is parked on the road \\
    \midrule
    \raisebox{-.65\height}{\includegraphics[width=\linewidth]{sample/26030.png}}
    & \plotspec{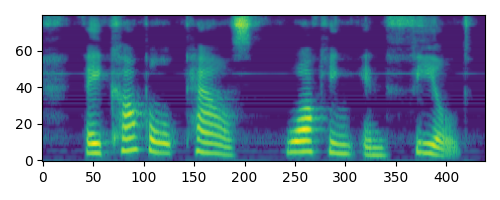} a couple of cows walking in a field
    & \plotspec{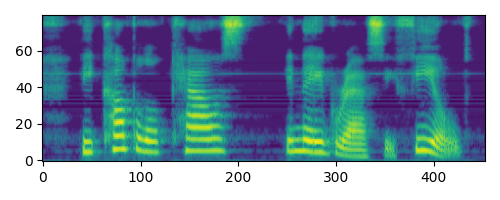} a couple of cows in a grassy field
    & \plotspec{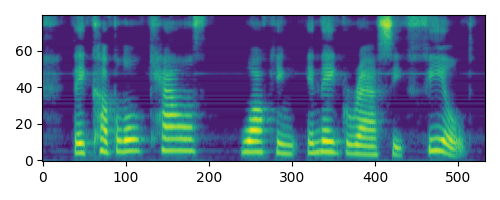} a couple of cows walking in a grassy field \\
    \midrule
    \raisebox{-.65\height}{\includegraphics[width=\linewidth]{sample/326.png}}
    & \plotspec{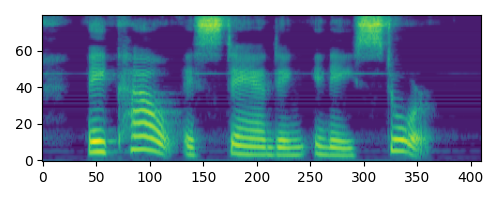} a cow is standing in a store
    & \plotspec{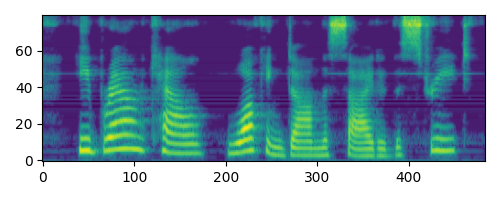} a brown cow walking down the side of a street
    & \plotspec{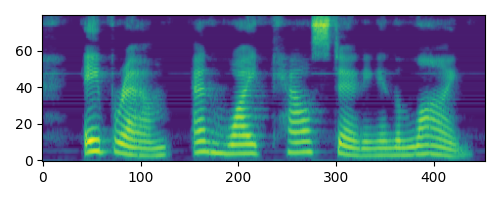} a brown and white cow standing in a line \\
    \midrule
    \raisebox{-.65\height}{\includegraphics[width=\linewidth]{sample/8402.png}}
    & \plotspec{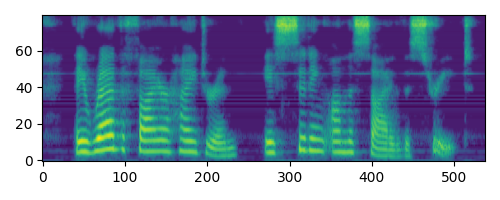} a red fire hydrant is sitting on the side of the street
    & \plotspec{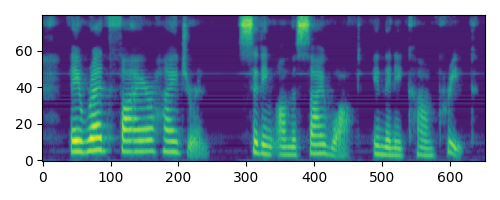} a red fire hydrant sitting on a sidewalk in a concrete
    & \plotspec{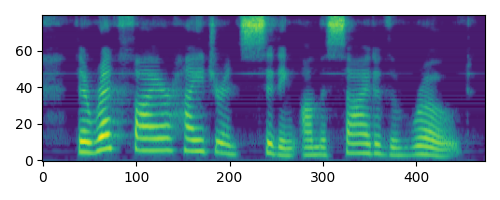} a red fire hydrant sitting on the side of a road \\
    \midrule
    \raisebox{-.65\height}{\includegraphics[width=\linewidth]{sample/37069.png}}
    & \plotspec{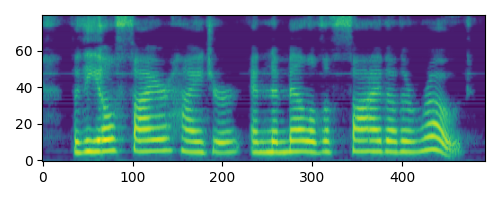} a yellow fire hydrant in the middle of the side of a road
    & \plotspec{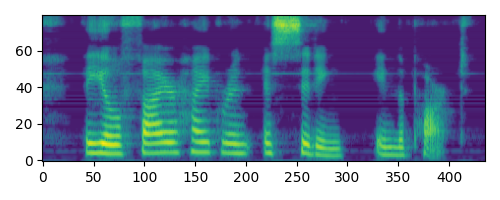} a yellow fire hydrant is sitting in the park
    & \plotspec{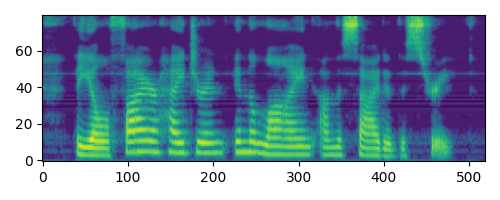} a yellow fire hydrant in a line on the side of a street \\
    \midrule
    \raisebox{-.65\height}{\includegraphics[width=\linewidth]{sample/8435.png}}
    & \plotspec{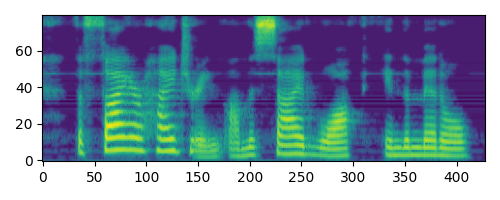} a fire hydrant on a sidewalk in the middle
    & \plotspec{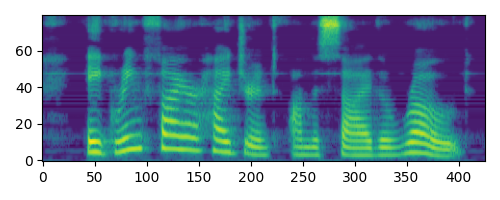} a green fire hydrant on the side of the road
    & \plotspec{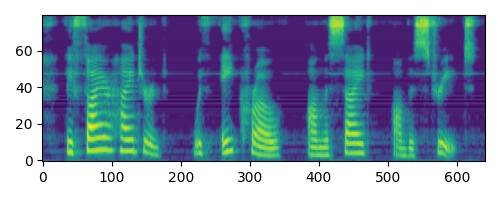} a fire hydrant with a curb on the side of the street \\
    \end{tabular}
    \caption{Samples. More at \url{https://wnhsu.github.io/image-to-speech-demo/2_vq3_sample_diversity_sat-ft_model}}
    \label{tab:i2s_demo_samp}
\end{table*}

\end{document}